\documentclass{article}
\usepackage[eabstract]{log_2025}

\usepackage{booktabs}
\usepackage{multirow}
\usepackage{amsfonts}
\usepackage{graphicx}
\usepackage{duckuments}

\usepackage{tikz}
\usepackage{amsmath}
\usepackage{courier}
\usepackage{helvet}

\usetikzlibrary{shapes,positioning,calc,decorations.text,matrix,arrows.meta,shapes.arrows}

\usepackage[numbers,compress,sort]{natbib}
\usepackage{subcaption}

\usepackage{listings}

\definecolor{codegreen}{rgb}{0,0.8,0}
\definecolor{codegray}{rgb}{0.3,0.3,0.3}
\definecolor{codepurple}{rgb}{0.8,0,0.8}
\definecolor{backcolour}{rgb}{0.95,0.95,0.92}

\lstdefinestyle{mystyle}{
    backgroundcolor=\color{backcolour},
    commentstyle=\color{codegreen},
    keywordstyle=\color{magenta},
    numberstyle=\tiny\color{codegray},
    stringstyle=\color{codepurple},
    basicstyle=\ttfamily\bfseries\fontsize{5}{6}\selectfont,
    breakatwhitespace=false,
    breaklines=true,
    captionpos=b,
    keepspaces=true,
    numbers=left,
    numbersep=5pt,
    showspaces=false,
    showstringspaces=false,
    showtabs=false,
    tabsize=2
}
\title{KNARsack: Teaching Neural Algorithmic Reasoners to Solve Pseudo-Polynomial Problems}

\author[Požgaj et al.]{%
  Stjepan Požgaj$^{1}$ \quad
  Dobrik Georgiev$^{2}$ \quad
  Marin Šilić$^{1}$ \quad
  Petar Veličković$^{3}$ \\
  $^1$University of Zagreb, Faculty of Electrical Engineering and Computing\\
  $^2$Graphcore\quad
  $^3$Google DeepMind\\
  \texttt{\{stjepan.pozgaj, marin.silic\}@fer.unizg.hr}\\
  \texttt{dobrikg@graphcore.ai} \quad
  \texttt{petarv@google.com}
}
\usepackage{amsmath,amsfonts,bm}

\def\eqref#1{equation~\ref{#1}}
\def\1{\bm{1}}

\def\rmE{{\mathbf{E}}}

\def\rmX{{\mathbf{X}}}

\def\vg{{\bm{g}}}

\def\vp{{\bm{p}}}

\def\vw{{\bm{w}}}

\DeclareMathAlphabet{\mathsfit}{\encodingdefault}{\sfdefault}{m}{sl}
\SetMathAlphabet{\mathsfit}{bold}{\encodingdefault}{\sfdefault}{bx}{n}

\def\gE{{\mathcal{E}}}

\def\gG{{\mathcal{G}}}

\def\gP{{\mathcal{P}}}

\def\gV{{\mathcal{V}}}

\def\sH{{\mathbb{H}}}
\def\sI{{\mathbb{I}}}

\def\sR{{\mathbb{R}}}

\begin{document}

\maketitle

\begin{abstract}
Neural algorithmic reasoning (NAR) is a growing field that aims to embed algorithmic logic into neural networks by imitating classical algorithms.
In this extended abstract, we detail our attempt to build a neural algorithmic reasoner that can solve Knapsack, a pseudo-polynomial problem bridging classical algorithms and combinatorial optimisation, but omitted in standard NAR benchmarks.
Our neural algorithmic reasoner is designed to closely follow the two-phase pipeline for the Knapsack problem, which involves first constructing the dynamic programming table and then reconstructing the solution from it.
The approach, which models intermediate states through dynamic programming supervision, achieves better generalization to larger problem instances than a direct-prediction baseline that attempts to select the optimal subset only from the problem inputs.
\end{abstract}

\section{Introduction}
Neural algorithmic reasoning \cite[NAR]{velivckovic2021neural} aims to connect the flexibility of machine learning with the structure and reliability of classical algorithms.
Rather than learning purely from input-output pairs, as in most standard machine learning pipelines, NAR models are usually trained under supervision at the algorithm's intermediate computation steps, enabling them to better generalize and reason algorithmically.
The CLRS-30 benchmark \cite{velivckovic2022clrs} was introduced to support this paradigm, covering 30 polynomial-time algorithms ranging from sorting to graph algorithms and dynamic programming.
However, many important combinatorial optimization problems -- including the Knapsack problem \citep{cormen2009introduction} -- fall outside its scope.

\begin{figure}
    \centering
    \includegraphics[width=0.95\textwidth]{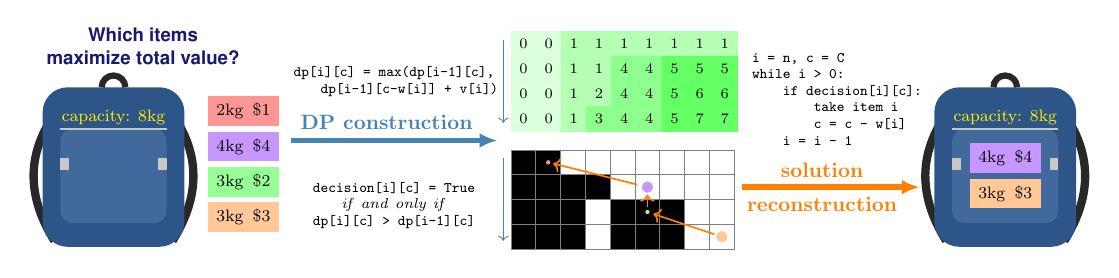}
    \caption{Visualization of the Knapsack problem and its dynamic programming solution. Left: input items with their weights and values. Middle: dynamic programming table and decision table. Right: optimal solution with selected items. \texttt{dp[i][c]} -- optimal value for the first $i$ items with capacity $c$.}
    \label{fig:knapsack}
\end{figure}

Knapsack is NP-hard. \cite{karp2009reducibility} CLRS-30 omits this class of problems, due to the impossibility of generating sufficiently many samples for them (unless NP=co-NP) \citep{yehuda2020machines}. However, for Knapsack, there exists a dynamic programming solution whose runtime depends on the numeric value of the capacity rather than its bit-length. This makes Knapsack a pseudo-polynomial problem\footnote{Informal term meaning the problem admits a pseudo-polynomial algorithm; formally, it is weakly NP-hard.}, lifting the implications of \citet{yehuda2020machines}, \emph{provided we do not consider extreme capacities}.
Despite its theoretical and practical significance, the problem has not been explored within the NAR framework.
Thus, we develop a neural algorithmic reasoner for Knapsack following a two-phase pipeline: dynamic programming table construction and solution reconstruction, as shown in \autoref{fig:knapsack}.
This approach is transferable to other pseudo-polynomial problems, as further discussed in \autoref{appendix:other_pseudopolynomial}.

In general, the contributions of our work can be summarised as follows:
1) we introduce a \textbf{two-step construction-reconstruction} NAR approach for pseudo-polynomial problems; 2) we outline several key design choices that enabled generalization to larger problem instances, namely \textbf{edge length encoding} and explicitly removing unnecessary input; 3) through results and appendices we provide extensive benchmarking and ablations of our approach giving insight into our architectural decisions.

\section{NAR for Knapsack}

\subsection{Problem Representation}
Given $n$ items, each with a weight $w_i$ and value $v_i$, and a capacity $C$, we aim to select a subset of the items such that the total weight does not exceed $C$, whilst maximizing value.
The problem is pseudo-polynomial if at least one of the weights or values are integers  \cite{wojtczak2018strong} -- we chose the weights to be integers. As current NAR models struggle with unbounded integers, we represent weights with a one-hot encoded input feature, limiting the weight to $w_{max}$. This is equivalent to lowering the capacity to $C:=\min(n\cdot w_{max}, C)$, bringing us in the ``non-extreme capacity'' case discussed in \S1.

For the problem representation, we use the CLRS-30 framework \cite{ibarz2022generalist}.
In it, any algorithm $A$ operates on graphs $\mathcal{G}=(\mathcal{V}, \mathcal{E})$ and the execution of $A$ is defined by probes, representing the set of node/edge/graph inputs ($\sI_\gV$/$\sI_\gE$/$\sI_\gG$), hints (${\sH}^{(t)}_\gV$/${\sH}^{(t)}_\gE$/${\sH}^{(t)}_\gG$), or outputs. The latent node/edge/graph features at timestep $t$  -- $\rmX^{(t)}\in \sR^{|\gV|\times d}$, $\rmE^{(t)}\in \sR^{|\gV|\times|\gV|\times d}$, and $\vg^{(t)}\in \sR^{d}$, where $d=128$ -- are obtained by encoding hints and inputs using encoders (various functions $f$ below) based on their specific types:
\begin{align*}
    \rmX^{(t)} = f^A_n(\sI_\gV) + \tilde{f}^A_n({\sH}^{(t)}_\gV)\qquad
    \rmE^{(t)} &= f^A_e(\sI_\gE) + \tilde{f}^A_e({\sH}^{(t)}_\gE)\qquad
    \vg^{(t)} = f^A_g(\sI_\gG) + \tilde{f}^A_g({\sH}^{(t)}_\gG)
\end{align*}
After encoding, the features are passed through a processor network and next step hints ($\sH^{(t+1)}_\gV$/$\sH^{(t+1)}_\gE$/$\sH^{(t+1)}_\gG$) are predicted from the latent embeddings. At inference the model uses its own \emph{soft} hint predictions \citep[p.5]{ibarz2022generalist}. For processor's architecture and losses we follow the CLRS-30 framework, unless otherwise stated, but we deviate on how we predict final outputs -- see below.

\subsection{Pseudo-Polynomial Modeling}
We split our training into training the model that constructs the DP table, and training the model to reconstruct the solution based on the input and the DP table generated by the first.
We initially attempted a unified approach but found it too unstable to train, resulting in poor inference performance.

\paragraph{Dynamic Programming Table Construction}
The dynamic programming state $\mathrm{dp}_{i, c}$ represents the maximum value achievable with the first $i$ items and a knapsack capacity of $c$. A $\mathrm{decision}_{i, c}$ variable indicates whether the $i$-th item is included in the optimal solution $\mathrm{dp}_{i, c}$, as shown in \autoref{fig:knapsack}.
As the DP table is constructed row by row, we chose $\mathcal{V}=\{u_0, \dots u_C\}$ for each $t\in \{1,\dots, n\}$. At timestep $t$ we set $\sI_\gG =\{w_t, v_t\}$. Then the model predicts $\sH^{(t)}_\gV=\{\mathrm{dp}_{t, :}, \mathrm{decision}_{t, :}\}$. The decision probability table is obtained by stacking the decision row predictions $\mathrm{decision}_{t,:}$ at each step.

\begin{figure}[]
    \centering
    \includegraphics[width=0.95\textwidth]{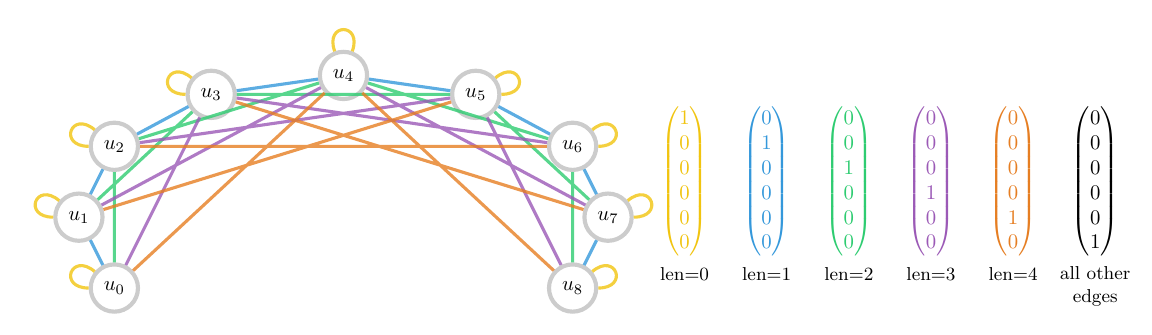}
    \caption{Categorical edge length encoding for the NAR construction model.
    Although we use $M=10$ in the implementation, here, due to visualization constraints, the case with $M=6$ is shown.}
    \label{fig:edge_positional_encoding_example}
\end{figure}

\begin{figure}
    \centering
    \begin{minipage}{0.27\textwidth}
        \centering
        \includegraphics[width=\textwidth]{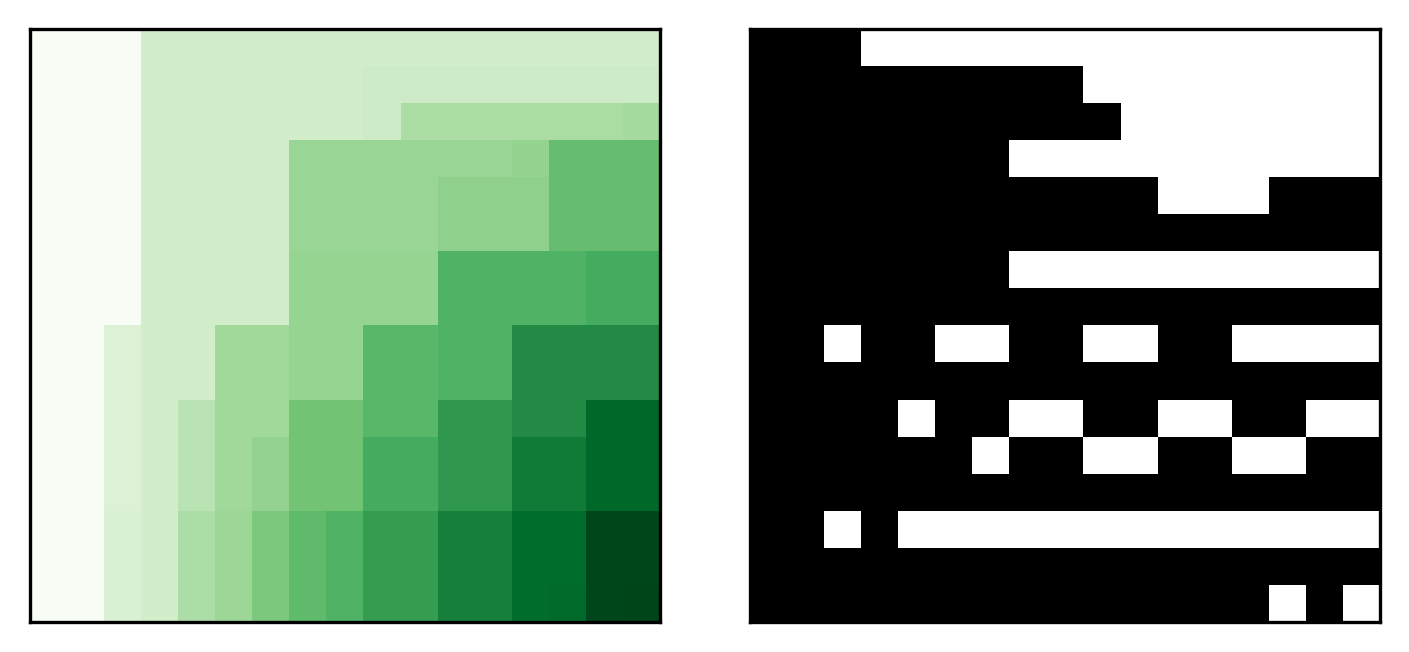}
        \subcaption{True DP tables}
    \end{minipage}\hspace{0.08\textwidth}
    \begin{minipage}{0.27\textwidth}
        \centering
        \includegraphics[width=\textwidth]{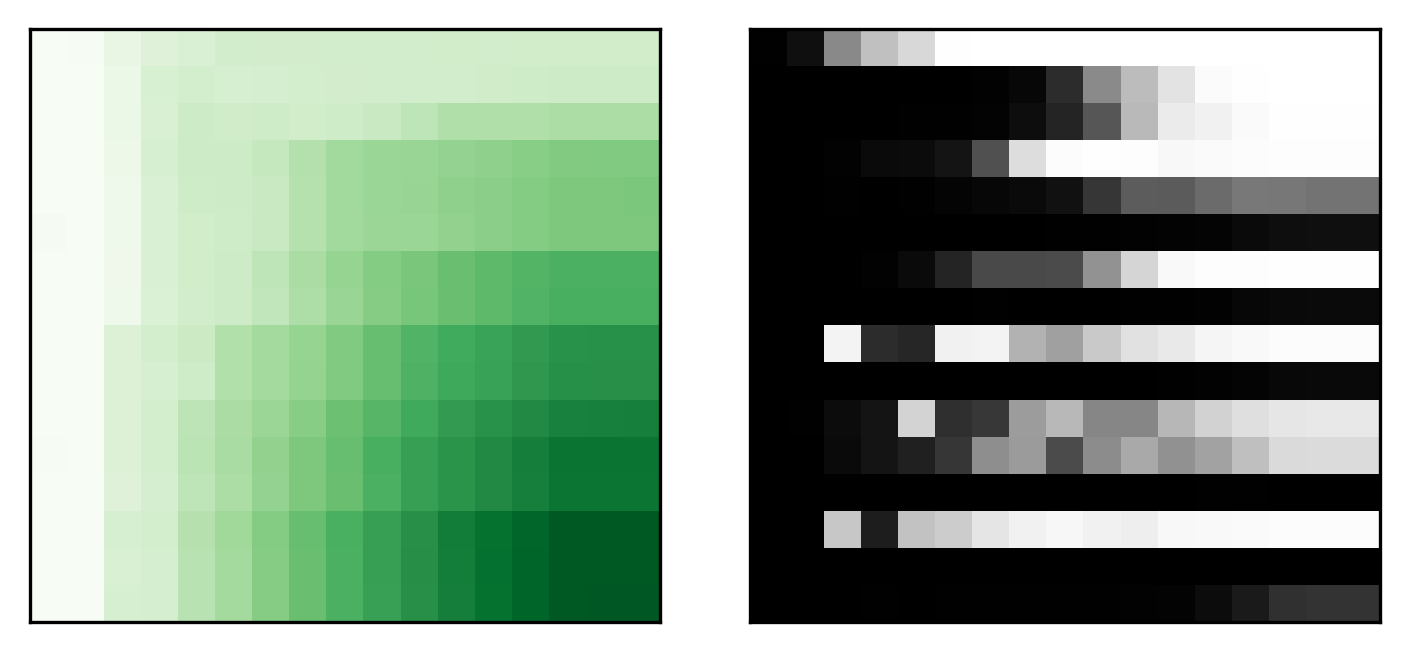}
        \subcaption{No edge length encoding}
    \end{minipage}\hspace{0.08\textwidth}
    \begin{minipage}{0.27\textwidth}
        \centering
        \includegraphics[width=\textwidth]{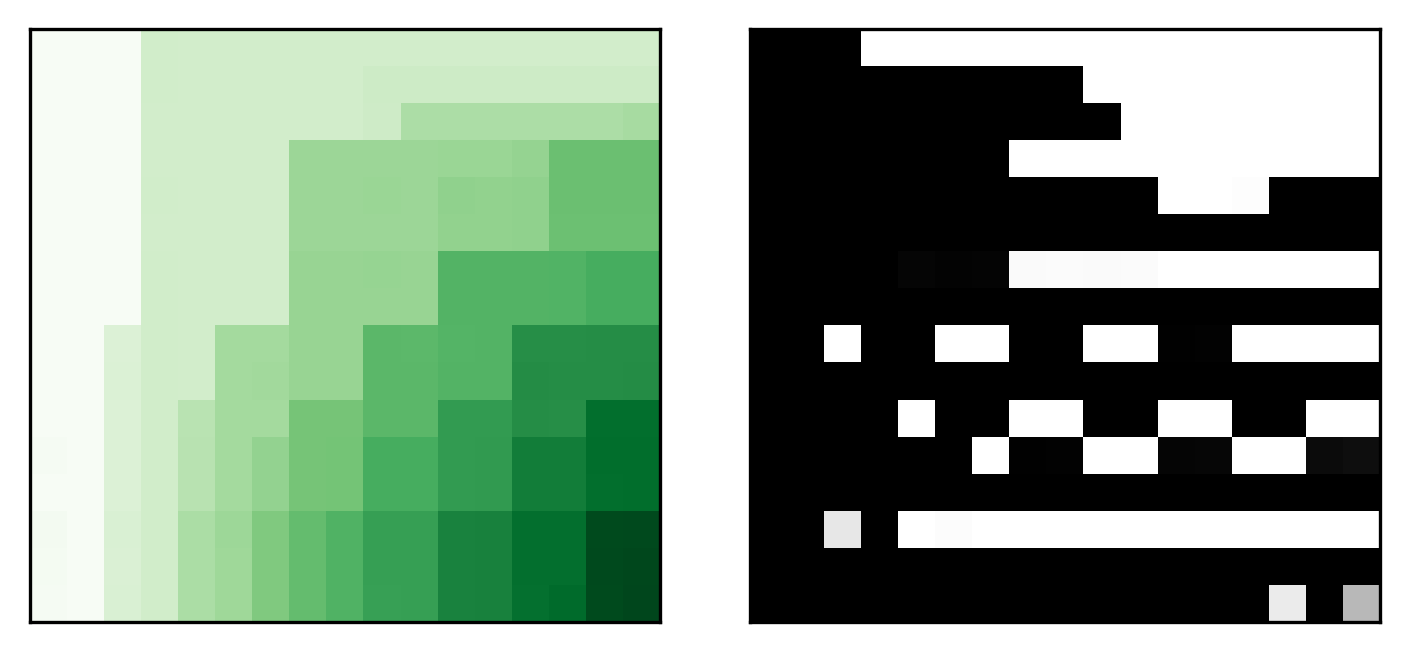}
        \subcaption{With edge length encoding}
    \end{minipage}
    \caption{Comparison of constructed DP tables for $n=16$, $C=16$ (in-distribution).
		Using categorical edge length encoding sharpens the predicted tables and enables reasoning.
	}
    \label{fig:edge_positional_encoding}
\end{figure}

\emph{Edge length encoding} was a crucial ingredient for successful training of the NAR constructor:
\begin{align*}
  \sI_\gE = \sI_\gE \cup \{\mathrm{EL}(\gE)\}  \qquad \mathrm{EL}(\mathcal{E})_{ijm}=\delta_{m,min(|i-j|, M-1)} \qquad \mathrm{EL}: \gP(\gE) \to\{0,1\}^{|\gV|\times |\gV| \times M}
\end{align*}
where  $\delta$ is the Kronecker delta function and $M=10$ is the cutoff. Such categorical encoding gives the model the ability to choose the correct past states that influence the current one.
Edge length encoding is illustrated in \autoref{fig:edge_positional_encoding_example}.
\autoref{fig:edge_positional_encoding} shows the impact of EL on an in-distribution problem instance in comparison to standard random node positional encoding.
For more details, see \autoref{appendix:edge_positional_encoding}.

We additionally observe that the CLRS-30's processors struggle to handle the Knapsack DP scalar values larger than those seen in training.
We propose to mitigate this issue by adopting a \emph{homogeneous processor} \cite{tang2020towards}, which enforces the model to be invariant to the scale of the item values.
This substantially improves the value generalization performance of the model on instances larger than those seen in training.
\autoref{appendix:homogeneity} provides more results and a discussion.

\paragraph{Solution Reconstruction}
Based on problem input and the predicted $\mathrm{decision}$ table, the solution is reconstructed by iteratively selecting the items that are part of the optimal set, as illustrated in \autoref{fig:knapsack}.
Unlike in the construction phase, in order to encode information about all items and the entire $\mathrm{decision}$ table, $\gV=\{u_0,\dots,u_C\}\cup\{u_{C+1},\dots _{C+n}\}$ where the second set are the item nodes.
\begin{align*}
\sI_\gV=\{\vp,
\vw, {\scriptsize[\,\underbrace{0, \dots, 0}_{C+1}, \;\underbrace{1, \dots, 1}_{n}\,]}\} \qquad {\sI_\gE}=\bigcup_{(c,i)\in\gE}\{\mathrm{decision}_{ic}\} \ \ \cup \bigcup_{(c, c') \in \gE}\{EL(\gE)_{cc'}\}
\end{align*}
where $\vp$ and $\vw$ are the vectors of positional encodings and weights (capacity nodes are padded with 0), $0\leq c, c' \leq C$ and $C+1\leq i\leq C+n$ and  $\sI_\gG=\emptyset$.
To model the reconstruction process, the hints provide information on the item currently under consideration, the total remaining capacity, and the set of items selected up to that point. For each item, the model outputs the probability that it belongs to the solution.
Due to the space constraints, details can be found in \autoref{appendix:knar_reconstruction}.

\begin{figure}
    \centering
    \includegraphics[width=0.95\textwidth]{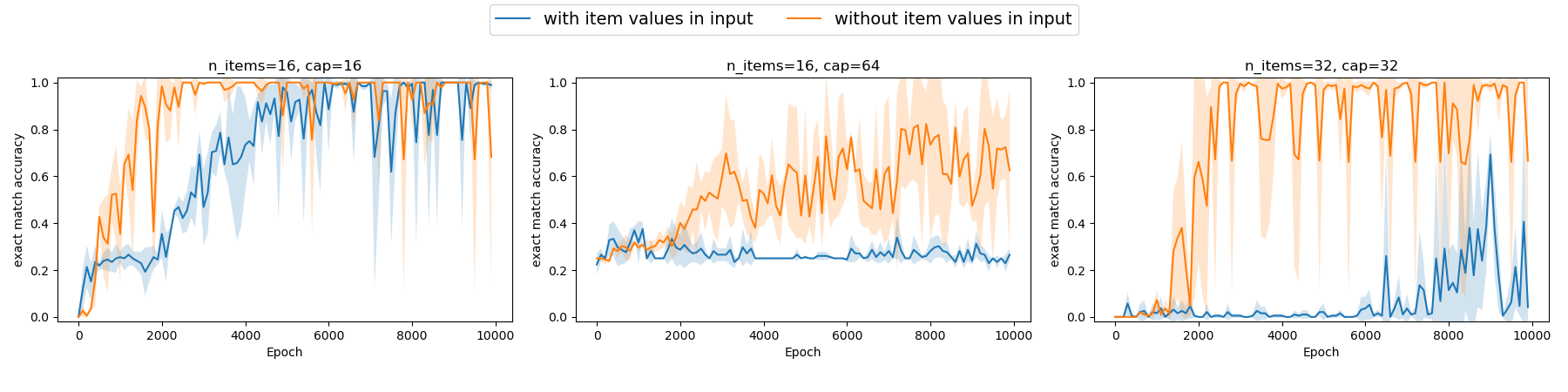}
    \caption{Comparison of NAR reconstruction performance with and without item values in the input, given the true $\mathrm{decision}$ table during both training and inference.
    Having the items' values in the input prevents the NAR reconstruction model from generalizing to larger instances.}
    \label{fig:first_3_with_vs_without_item_values}
\end{figure}

To reconstruct the solution from the $\mathrm{decision}$ table, it is sufficient to have information about the item weights.
During reconstruction, as shown in \autoref{fig:first_3_with_vs_without_item_values}, \emph{having item values leads to worse generalization}, as the model is trying to find a shortcut to the solution, without actually learning to navigate the $\mathrm{decision}$ table.
We hypothesise that this is the reason why the joint model approach did not work.
We also note that \emph{having two alternating steps}, one for decision and one for moving the pointer works better than combining the two.
Further details are provided in \autoref{fig:one_vs_two_recon_hints}, \autoref{appendix:knar_reconstruction}.

\paragraph{Deterministic Reconstruction}
An alternative would be to perform reconstruction based on the $\mathrm{decision}$ probability table, as it allows for end-to-end differentiability.
In \autoref{appendix:deterministic_reconstruction}, we describe in more detail such a differentiable deterministic reconstruction.
Sadly, our initial experiments in training an NAR construction, coupled with deterministic reconstruction, were unstable.
Nevertheless, in the evaluation, we include as a baseline a model that, during the inference phase, performs deterministic reconstruction based on the $\mathrm{decision}$ table generated by NAR construction model.

\section{Evaluation}
\paragraph{Experimental Setup}
Our experiments are implemented on top of CLRS-30 \cite{ibarz2022generalist} (3 seeds for std).
Hyperparameters (learning rate, optimiser, etc.) are inherited, including the use of Triplet-GMPNN processor \cite{dudzik2022graph} as well as the standard node positional encoding.
Differently from CLRS-30, we use the model from the final epoch for testing, as it was recommended in \cite{mahdavi2022towards}.
Furthermore, during the construction training phase, we perform 30,000 epochs, as we observed that the model does not stabilize within the first 10,000 epochs.

The model is trained on samples where $n \leq 16$ and $C \leq 16$ (in-distribution).
For out-of-distribution experiments, we consider different values of $n$ and $C$ (up to $n=64$ and $C=64$).
Each configuration consists of 64 samples.
Item weights are uniformly sampled from $\{1, 2, \ldots, 8\}$, and item values are uniformly sampled from the interval $[0, 1]$.

Following CLRS-30, during testing, we report the micro-F1 score. We also add
\textit{exact-match accuracy} -- the fraction of samples whose discretized solution, obtained by greedily selecting the highest-probability item until capacity is reached, exactly matches the ground truth.

\paragraph{Models and Baselines}
Alongside the models with regular processors (\textbf{reg.\ c.\ + reg.\ r.}), and a homogeneous processor during construction (\textbf{homo.\ c.\ + reg.\ r.}), we evaluate a \textbf{no-hint baseline}, which directly predicts the optimal subset from the input. It is implemented using the \textit{no-hint} mode in CLRS-30 with $2n$ message-passing steps, and includes scalar features for item weights and capacity. We additionally zero-shot evaluate the homogeneous construction processor with the deterministic reconstruction baseline (\textbf{homo.\ c.\ + det.\ r.}), despite the fact they were not trained together.
We also compare our models to Tropical Attention \citep{hashemi2025tropical}, a recent direct-prediction model for combinatorial optimization problems based on tropical projective geometry and polyhedral geometric inductive bias.
To make the comparison, we adjusted the sampler and the training and test sizes to match ours, see \autoref{appendix:adapting_tropical_attention}.
For more information on this and other relevant works, see \autoref{appendix:related_work}.

\subsection{Results}

\def\regreg{(\textbf{reg.\ c.\ + reg.\ r.})}
\def\homoreg{(\textbf{homo.\ c.\ + reg.\ r.})}
\def\homodet{(\textbf{homo.\ c.\ + det.\ r.})}

\begin{table}\centering
\caption{\textcolor{Orange}{In-} and \textcolor{DarkOrchid}{out-} of distribution performance for different models. Best results are \textbf{boldfaced}, but we separate \textcolor{Blue}{models using an oracle in their execution}. Best viewed on screen.}\label{tab:main_results}
\resizebox{\linewidth}{!}{%
\begin{tabular}{ll
c@{\hskip 5pt}c
c@{\hskip 5pt}c
c@{\hskip 5pt}c
c@{\hskip 5pt}c
c@{\hskip 5pt}c}
\toprule
\multirow{2.5}{*}{$n$} & \multirow{2.5}{*}{$C$} &
\multicolumn{2}{c}{no-hint baseline} &
\multicolumn{2}{c}{reg. c. + reg. r.} &
\multicolumn{2}{c}{homo. c. + reg. r.} &
\multicolumn{2}{c}{homo. c. + det. r.} &
\multicolumn{2}{c}{Tropical Attention \citep{hashemi2025tropical}} \\
\cmidrule(lr){3-4} \cmidrule(lr){5-6} \cmidrule(lr){7-8} \cmidrule(lr){9-10} \cmidrule(lr){11-12}
& & micro-F1 & exact & micro-F1 & exact & micro-F1 & exact & micro-F1 & exact & micro-F1 & exact \\
\midrule
\textcolor{Orange}{$16$} & \textcolor{Orange}{$16$} &
$0.889_{\scriptscriptstyle\pm0.017}$ & $0.385_{\scriptscriptstyle\pm0.101}$ &
$\mathbf{0.989}_{\scriptscriptstyle\pm0.004}$ & $\mathbf{0.906}_{\scriptscriptstyle\pm0.031}$ &
$0.985_{\scriptscriptstyle\pm0.007}$ & $0.901_{\scriptscriptstyle\pm0.033}$ &
$0.975_{\scriptscriptstyle\pm0.004}$ & $0.891_{\scriptscriptstyle\pm0.013}$ &
$0.909_{\scriptscriptstyle\pm0.013}$ & $0.381_{\scriptscriptstyle\pm0.005}$ \\
\midrule
\textcolor{DarkOrchid}{$16$} & \textcolor{DarkOrchid}{$64$} &
$\mathbf{0.943}_{\scriptscriptstyle\pm0.000}$ & $0.250_{\scriptscriptstyle\pm0.000}$ &
$0.785_{\scriptscriptstyle\pm0.063}$ & $0.438_{\scriptscriptstyle\pm0.095}$ &
$0.917_{\scriptscriptstyle\pm0.017}$ & $\mathbf{0.495}_{\scriptscriptstyle\pm0.115}$ &
$\textcolor{Blue}{\mathbf{0.957}}_{\scriptscriptstyle\pm0.012}$ & $\textcolor{Blue}{\mathbf{0.578}}_{\scriptscriptstyle\pm0.109}$ &
$0.892_{\scriptscriptstyle\pm0.011}$ & $0.493_{\scriptscriptstyle\pm0.027}$ \\
\textcolor{DarkOrchid}{$32$} & \textcolor{DarkOrchid}{$32$} &
$0.656_{\scriptscriptstyle\pm0.122}$ & $0.000_{\scriptscriptstyle\pm0.000}$ &
$0.854_{\scriptscriptstyle\pm0.015}$ & $0.177_{\scriptscriptstyle\pm0.039}$ &
$\mathbf{0.925}_{\scriptscriptstyle\pm0.040}$ & $\mathbf{0.510}_{\scriptscriptstyle\pm0.173}$ &
$\textcolor{Blue}{\mathbf{0.933}}_{\scriptscriptstyle\pm0.034}$ & $\textcolor{Blue}{\mathbf{0.521}}_{\scriptscriptstyle\pm0.128}$ &
$0.923_{\scriptscriptstyle\pm0.012}$ & $0.186_{\scriptscriptstyle\pm0.078}$ \\
\textcolor{DarkOrchid}{$64$} & \textcolor{DarkOrchid}{$16$} &
$0.835_{\scriptscriptstyle\pm0.030}$ & $0.031_{\scriptscriptstyle\pm0.031}$ &
$0.824_{\scriptscriptstyle\pm0.055}$ & $0.135_{\scriptscriptstyle\pm0.127}$ &
$\mathbf{0.939}_{\scriptscriptstyle\pm0.011}$ & $\mathbf{0.615}_{\scriptscriptstyle\pm0.110}$ &
$\textcolor{Blue}{\mathbf{0.949}}_{\scriptscriptstyle\pm0.015}$ & $\textcolor{Blue}{\mathbf{0.635}}_{\scriptscriptstyle\pm0.102}$ &
$0.822_{\scriptscriptstyle\pm0.024}$ & $0.043_{\scriptscriptstyle\pm0.047}$ \\
\textcolor{DarkOrchid}{$64$} & \textcolor{DarkOrchid}{$64$} &
$0.543_{\scriptscriptstyle\pm0.000}$ & $0.000_{\scriptscriptstyle\pm0.000}$ &
$0.448_{\scriptscriptstyle\pm0.084}$ & $0.000_{\scriptscriptstyle\pm0.000}$ &
$\mathbf{0.668}_{\scriptscriptstyle\pm0.087}$ & $0.000_{\scriptscriptstyle\pm0.000}$ &
$\textcolor{Blue}{\mathbf{0.770}}_{\scriptscriptstyle\pm0.115}$ & $0.000_{\scriptscriptstyle\pm0.000}$ &
$0.374_{\scriptscriptstyle\pm0.000}$ & $0.000_{\scriptscriptstyle\pm0.000}$ \\
\bottomrule
\end{tabular}
}
\end{table}

Results are presented in \autoref{tab:main_results}. Our first observation is that the combination of two regular processors does not obtain higher F1-scores than the baseline and even falls short on the most extreme OOD (64, 64) test set. Despite that, exact-match accuracy for \regreg\ is often significantly higher than the baseline. The results suggest that: 1) the construction-reconstruction pipeline works differently than the baseline and 2) node-level metrics should often be combined with instance-level ones \citep{mahdavi2022towards}.

The second result we want to highlight is the increase in performance when using a homogeneous NAR constructor. In the larger OOD regimes \homoreg\ significantly outperforms the baseline F1 scores and achieves best exact-match accuracy from the fully-neural models. According to our observations, even if the input value scales remain fixed across test distributions, homogeneity could be one of the requirements when modelling pseudo-polynomial DP problems, since the magnitude of the solution increases with size (\autoref{appendix:homogeneity}).

The best results were achieved with the \homodet\ combination. Although we did observe improvements over \homoreg\ in both metrics, results were comparable. Our conclusions are: 1) \textbf{perfect} DP construction execution might be necessary and 2) our neural reconstruction behaves similarly to the deterministic one.

Our best fully neural model \homoreg\ achieves better performance than Tropical Attention in all OOD settings, but it is important to interpret these results in the context of the differences between approaches.
We note that Tropical Attention's primary aim, like ours, is not to compete with traditional or neural Knapsack solvers, but rather to demonstrate the effectiveness of their proposed architecture.
We further hypothesize that combining Tropical Attention's insights with ours -- which are largely orthogonal -- could yield even stronger performance.

\subsection{Properties of Generated Scalar DP Tables}
In \autoref{tab:homogeneity_construction}, \autoref{appendix:homogeneity}, we report qualities of the generated $\mathrm{decision}$ tables, but -- aside from a few visual examples -- we do not quantify the quality of the predicted scalar DP tables.
Because mean squared error is not very interpretable for this purpose, we perform an analysis based on three statistics that correspond to the fundamental properties of the Knapsack DP table. To avoid issues with numerical precision of predicted values, we use $\varepsilon = 0.01$, which is sufficiently precise for gaining insights into the constructed DP tables, given that item values are sampled from $[0, 1]$. These statistics measure the proportion of DP table cells for which the specified constraints hold.

\begin{table}\centering
\caption{Proportions of DP table entries satisfying three fundamental properties under $\varepsilon{=}0.01$: (1) item-wise monotonicity: $\mathrm{dp}_{i, c} \geq \mathrm{dp}_{i-1, c} - \varepsilon$; (2) capacity-wise monotonicity: $\mathrm{dp}_{i, c} \geq \mathrm{dp}_{i, c-1} - \varepsilon$; (3) optimal substructure: $\mathrm{dp}_{i, c} \geq \max(\mathrm{dp}_{i-1, c}, \mathrm{dp}_{i-1, c-w_i} + v_i) - \varepsilon$.}
\label{tab:dp_properties_combined}
\centering
\begin{tabular}{ll cc cc cc}
\toprule
\multirow{2.5}{*}{$n$} & \multirow{2.5}{*}{$C$} &
\multicolumn{2}{c}{item-wise mon.} &
\multicolumn{2}{c}{capacity-wise mon.} &
\multicolumn{2}{c}{optimal substructure} \\
\cmidrule(lr){3-4} \cmidrule(lr){5-6} \cmidrule(lr){7-8}
& & reg. c. & homo. c. & reg. c. & homo. c. & reg. c. & homo. c. \\
\midrule
\textcolor{Orange}{$16$} & \textcolor{Orange}{$16$} & $0.893_{\scriptscriptstyle\pm0.032}$ & $\mathbf{0.974}_{\scriptscriptstyle\pm0.009}$ & $\mathbf{0.951}_{\scriptscriptstyle\pm0.027}$ & $0.945_{\scriptscriptstyle\pm0.036}$ & $0.700_{\scriptscriptstyle\pm0.057}$ & $\mathbf{0.948}_{\scriptscriptstyle\pm0.015}$ \\
\midrule
\textcolor{DarkOrchid}{$16$} & \textcolor{DarkOrchid}{$64$} & $0.915_{\scriptscriptstyle\pm0.049}$ & $\mathbf{0.988}_{\scriptscriptstyle\pm0.004}$ & $\mathbf{0.892}_{\scriptscriptstyle\pm0.035}$ & $0.799_{\scriptscriptstyle\pm0.062}$ & $0.377_{\scriptscriptstyle\pm0.070}$ & $\mathbf{0.677}_{\scriptscriptstyle\pm0.043}$ \\
\textcolor{DarkOrchid}{$64$} & \textcolor{DarkOrchid}{$16$} & $0.711_{\scriptscriptstyle\pm0.014}$ & $\mathbf{0.978}_{\scriptscriptstyle\pm0.003}$ & $0.975_{\scriptscriptstyle\pm0.010}$ & $\mathbf{0.981}_{\scriptscriptstyle\pm0.014}$ & $0.548_{\scriptscriptstyle\pm0.031}$ & $\mathbf{0.967}_{\scriptscriptstyle\pm0.006}$ \\
\textcolor{DarkOrchid}{$32$} & \textcolor{DarkOrchid}{$32$} & $0.773_{\scriptscriptstyle\pm0.028}$ & $\mathbf{0.983}_{\scriptscriptstyle\pm0.004}$ & $\mathbf{0.967}_{\scriptscriptstyle\pm0.009}$ & $0.967_{\scriptscriptstyle\pm0.018}$ & $0.447_{\scriptscriptstyle\pm0.060}$ & $\mathbf{0.901}_{\scriptscriptstyle\pm0.028}$ \\
\textcolor{DarkOrchid}{$64$} & \textcolor{DarkOrchid}{$64$} & $0.636_{\scriptscriptstyle\pm0.022}$ & $\mathbf{0.981}_{\scriptscriptstyle\pm0.005}$ & $\mathbf{0.860}_{\scriptscriptstyle\pm0.031}$ & $0.841_{\scriptscriptstyle\pm0.064}$ & $0.241_{\scriptscriptstyle\pm0.032}$ & $\mathbf{0.715}_{\scriptscriptstyle\pm0.036}$ \\
\bottomrule
\end{tabular}
\end{table}

The results in \autoref{tab:dp_properties_combined} show that the homogeneous processor model produces significantly better tables in terms of item-wise monotonicity and optimal substructure property, while capacity-wise monotonicity results are comparable.
The results suggest that capacity generalization presents a greater challenge than item count generalization when predicting DP scalars.
While the optimal substructure property was violated 30\% of the time for our largest OOD experiment, the 46\% improvement was sufficient to improve our downstream predictions.
Note that there exist other alternative approaches \cite{rodionovdiscrete}, that provably generalise to \textit{any} input size, but those mimic the algorithms too closely, trading the flexibility of neural networks, which is one of the reasons NAR exists.

\paragraph{Limitations and Future Work} In spite of the solid results, there are limitations remaining to be addressed: 1) \textbf{Oracle instability} -- we were unable to train \homodet, due to exploding/vanishing gradients. We hypothesise this is due to the algorithm chaining multiplications of probabilities and/or operating in scalar space (\autoref{appendix:deterministic_reconstruction}). 2) \textbf{Separated trainings} -- currently, the training of the construction model is completely decoupled from the training of the reconstruction one. This is not common for NAR models and is a topic of our future interests.
Looking ahead, we also plan to investigate how the observations made in this work, such as the use of edge length encoding, could be applied to problems from CLRS-30, shedding light on their potential beyond the current setting.

\newpage

\section*{Acknowledgements}
The authors would like to thank Federico Barbero (Google DeepMind), Simon Osindero (Google
DeepMind), and Ndidi Elue (Google DeepMind) for their valuable comments and suggestions on this work.
S.P. thanks Pietro Li{\`o} for hosting him in his research group at University of Cambridge during part of this work.

\bibliographystyle{unsrtnat}
\bibliography{reference}

\appendix
\section{Related Work}
\label{appendix:related_work}
Neural algorithmic reasoning aligns learned computation with classical algorithms, with the aim of obtaining trained neural models that can execute the target algorithm in the out-of-distribution regime. Intermediate-step supervision \citep{velickovic2020Neural,velivckovic2022clrs}, also utilised here, has been the initial approach to achieve such alignment. However, this alone is not sufficient and architectural modifications were necessary. Early analyses studied what neural networks can reason about and how they extrapolate \citep{xu2019can,xu2021how} -- on an informal level their findings state that neural networks extrapolate well if the algorithm can be separated into subfunctions and each of them is easily learnable by a corresponding neural submodule. This, combined with the rise of standardised NAR benchmarks (CLRS-30 \citep{velivckovic2022clrs} and SALSA-CLRS \citep{minder2023salsaclrs}) gave rise to many subsequent works, proposing architectural alignment strategies such as homogeneity \citep{tang2020towards}, persistent message passing \citep{strathmann2021persistent}, relational/positional attention \citep{diao2022relational,luca2025positional}, and recurrent aggregators \citep{xu2024recurrent}. Related efforts explore structured memory for recursion and data structures \citep{jurss2024recursive,jain2023neural}, while others investigate reducing or re-structuring hints \citep{rodionov2023neural,mahdavi2022towards,georgiev2024deep,bevilacqua2023neural,li2024open}.

Graph neural networks \citep{VELICKOVIC2023102538}, the underlying architecture behind the NAR processors, are theoretically aligned to dynamic programming algorithms \citep{dudzik2022graph}.
However, when NAR is benchmarked, most studies address polynomial-time DPs, and the $0$--$1$ Knapsack algorithm, being pseudo-polynomial (weakly NP-hard), was deliberately omitted from CLRS-30.
While some works have addressed NP-hard problems \citep{georgiev2024neural, heprimal}, none have focused specifically on pseudo-polynomial problems.
Outside of the field of NAR, there exist DP-inspired neural networks \citep{hertrich2023provably} that provably solve Knapsack, but their weights are hardcoded, rather than learned from data, as is the case in our work.
Due to the nature of our problem, classical neural combinatorial optimisation (NCO) approaches could also be applied.
Some possible approaches are \citet{bello2016neural,kwon2020pomo,afshar2020state,zhang2025reinforcement}, but these are fundamentally different from our approach.
Moreover, these approaches represent item values and weights as real numbers and use fixed capacity.
Additionally, they do not consider out-of-distribution settings at all -- the instance size is fixed and ranges up to several hundred items.
It is worth noting that the idea of combining NAR and RL is very recent \citep{schutz2025tackling}, and there are no approaches that would tackle pseudo-polynomial algorithms in this manner.

From the NCO approaches, our work is most conceptually similar to supervised methods that solve Knapsack by learning the input-output mapping directly, without following the algorithm's intermediate steps \citep{drakulic2023bq, nomer2020neural, hashemi2025tropical}.
One supervised approach \citep{drakulic2023bq}, which uses the same sampler as \citet{kwon2020pomo}, does consider OOD, but trains on sizes of $100$ and tests on $200$, $500$, and $1000$, which is not comparable to our setting.
Another supervised learning approach for solving Knapsack \citep{nomer2020neural} does not have available source code and reproducibility is not at the required level.
The only work we found that is comparable to ours in terms of the training and testing scales is Tropical Attention \citep{hashemi2025tropical}. Tropical Attention \citep{hashemi2025tropical} introduces a solution based on tropical projective geometry and polyhedral geometric inductive bias. The Tropical Transformers enhance OOD performance in both length and value generalization on several combinatorial optimization problems.
In the Knapsack experiments, the model is trained for $n=8$ and tested for length generalization to $n=64$. Capacity generalization is not performed -- capacity is integer from $10$ to $20$, in both cases.
The model has much shorter training time, primarily due to the lightweight implementation and the fact that it does not use hints, but rather performs supervised direct-prediction of outputs based on inputs. The source code for Tropical Attention is available, so we adapted the sampler and training and test sizes to compare it with our approach.

\section{Other Pseudo-Polynomial Problems}
\label{appendix:other_pseudopolynomial}

\begin{table}[h]
    \centering
    \caption{\textcolor{Orange}{In-} and \textcolor{DarkOrchid}{out-} of distribution performance for the Subset Sum problem.}
    \label{tab:subset_sum}
    \begin{tabular}{cccccc}
        \toprule
        \multirow{2.5}{*}{$n_{\text{numbers}}$} & \multirow{2.5}{*}{$\text{target}$} &
        \multicolumn{2}{c}{no-hint baseline} &
        \multicolumn{2}{c}{homo. c. + reg. r.} \\
        \cmidrule(lr){3-4} \cmidrule(lr){5-6}
        & & micro-F1 & exact & micro-F1 & exact \\
        \midrule
        \textcolor{Orange}{$16$} & \textcolor{Orange}{$16$} &
        $0.770_{\scriptscriptstyle\pm0.014}$ & $0.354_{\scriptscriptstyle\pm0.024}$ &
        $\mathbf{1.000}_{\scriptscriptstyle\pm0.000}$ & $\mathbf{1.000}_{\scriptscriptstyle\pm0.000}$ \\
        \midrule
        \textcolor{DarkOrchid}{$16$} & \textcolor{DarkOrchid}{$64$} &
        $\mathbf{0.886}_{\scriptscriptstyle\pm0.000}$ & $\mathbf{0.266}_{\scriptscriptstyle\pm0.000}$ &
        $0.621_{\scriptscriptstyle\pm0.121}$ & $0.255_{\scriptscriptstyle\pm0.018}$ \\
        \textcolor{DarkOrchid}{$32$} & \textcolor{DarkOrchid}{$32$} &
        $0.406_{\scriptscriptstyle\pm0.094}$ & $0.365_{\scriptscriptstyle\pm0.036}$ &
        $\mathbf{0.854}_{\scriptscriptstyle\pm0.135}$ & $\mathbf{0.542}_{\scriptscriptstyle\pm0.423}$ \\
        \textcolor{DarkOrchid}{$64$} & \textcolor{DarkOrchid}{$16$} &
        $0.666_{\scriptscriptstyle\pm0.056}$ & $0.380_{\scriptscriptstyle\pm0.055}$ &
        $\mathbf{0.993}_{\scriptscriptstyle\pm0.008}$ & $\mathbf{0.990}_{\scriptscriptstyle\pm0.009}$ \\
        \textcolor{DarkOrchid}{$64$} & \textcolor{DarkOrchid}{$64$} &
        $0.353_{\scriptscriptstyle\pm0.000}$ & $\mathbf{0.141}_{\scriptscriptstyle\pm0.000}$ &
        $\mathbf{0.530}_{\scriptscriptstyle\pm0.106}$ & $0.094_{\scriptscriptstyle\pm0.041}$ \\
        \bottomrule
    \end{tabular}
\end{table}

\begin{table}[h]
    \centering
    \caption{\textcolor{Orange}{In-} and \textcolor{DarkOrchid}{out-} of distribution performance for the Partition problem.}
    \label{tab:partition_problem}
    \begin{tabular}{cccccc}
        \toprule
        \multirow{2.5}{*}{$n_{\text{numbers}}$} & \multirow{2.5}{*}{$\text{sum}$} &
        \multicolumn{2}{c}{no-hint baseline} &
        \multicolumn{2}{c}{homo. c. + reg. r.} \\
        \cmidrule(lr){3-4} \cmidrule(lr){5-6}
        & & micro-F1 & exact & micro-F1 & exact \\
        \midrule
        \textcolor{Orange}{$8$} & \textcolor{Orange}{$32$} &
        $0.813_{\scriptscriptstyle\pm0.007}$ & $0.406_{\scriptscriptstyle\pm0.027}$ &
        $\mathbf{0.996}_{\scriptscriptstyle\pm0.007}$ & $\mathbf{0.990}_{\scriptscriptstyle\pm0.018}$ \\
        \midrule
        \textcolor{DarkOrchid}{$16$} & \textcolor{DarkOrchid}{$64$} &
        $0.093_{\scriptscriptstyle\pm0.154}$ & $\mathbf{0.063}_{\scriptscriptstyle\pm0.000}$ &
        $\mathbf{0.698}_{\scriptscriptstyle\pm0.027}$ & $0.042_{\scriptscriptstyle\pm0.036}$ \\
        \textcolor{DarkOrchid}{$24$} & \textcolor{DarkOrchid}{$96$} &
        $0.037_{\scriptscriptstyle\pm0.064}$ & $0.000_{\scriptscriptstyle\pm0.000}$ &
        $\mathbf{0.578}_{\scriptscriptstyle\pm0.088}$ & $0.000_{\scriptscriptstyle\pm0.000}$ \\
        \textcolor{DarkOrchid}{$32$} & \textcolor{DarkOrchid}{$128$} &
        $0.030_{\scriptscriptstyle\pm0.052}$ & $0.000_{\scriptscriptstyle\pm0.000}$ &
        $\mathbf{0.572}_{\scriptscriptstyle\pm0.123}$ & $0.000_{\scriptscriptstyle\pm0.000}$ \\
        \bottomrule
    \end{tabular}
\end{table}
While this work focuses exclusively on the Knapsack problem, the underlying approach is readily transferable to other pseudo-polynomial problems.
Both the Subset Sum and Partition problem can be directly reduced to the Knapsack formulation, which allowed us to apply our method with only minimal modifications to the sampler. The results for these problems are reported in \autoref{tab:subset_sum} and \autoref{tab:partition_problem}. Even without any problem-specific tuning, our model tended to generalize better than the baseline in most cases. Extending the approach to other pseudo-polynomial problems -- such as Minimum Coin Exchange or Rod Cutting -- would require adjustments to the problem representation. Nevertheless, the core methodology can be applied without substantial changes.

\section{Additional Details on the Experimental Setup} \label{appendix:adapting_tropical_attention}

\paragraph{Training and Testing Scales}
Training on sizes up to 16 with OOD testing on size 64 is the standard for the CLRS-30 benchmark \cite{velivckovic2022clrs} and is therefore followed by the vast majority of NAR works.
Since we ultimately aim to include pseudo-polynomial problems in the benchmark, this emerged as the most natural choice.
Given that in our reconstruction model the number of graph nodes equals $N+C+1$, training on $(n=32, C=32)$ is already not memory-feasible within the CLRS-30 framework.
If we consider node accuracy alone, $4\times$ generalization is gradually becoming obsolete as OOD test, e.g., for the Bellman-Ford algorithm.
However, from our baselines (the default models in CLRS-30), we can see that this is not the case for Knapsack.
Additionally, in our case we have two parameters that determine the problem size, while all CLRS-30 algorithms have only one parameter.
The size of our DP table is $N\times(C+1)$, so with our OOD experiments on parameters $(n=16, C=64)$, $(n=64, C=16)$, and $(n=32, C=32)$, we effectively test $4\times$ generalization as in CLRS-30, as the number of total elements in the table grows fourfold, and with the additional $(n=64, C=64)$ we test up to $16\times$ generalization.

\begin{figure}[]
    \centering
    \includegraphics[width=0.95\textwidth]{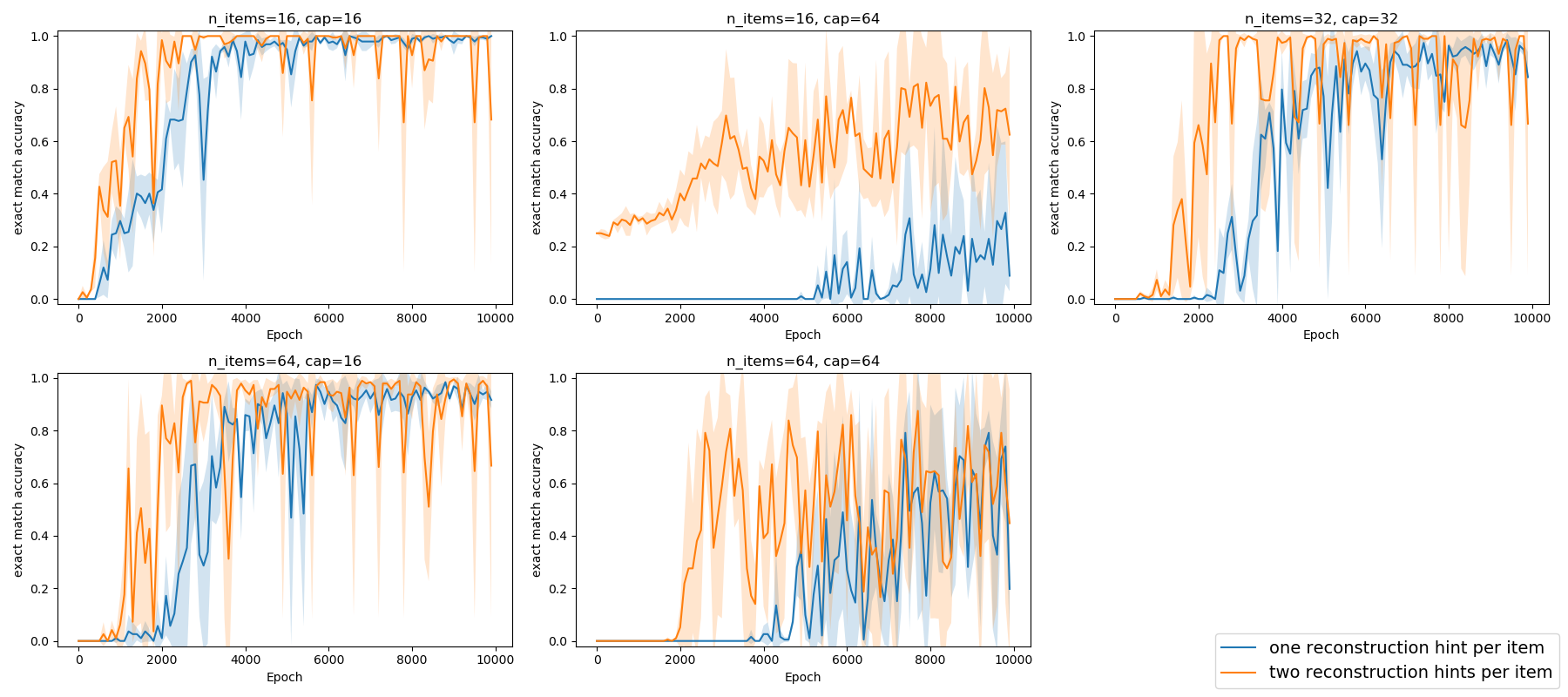}
    \caption{Comparison of the joint-step and split-step NAR reconstruction performances, given the true $\mathrm{decision}$ table during both training and inference.
    Splitting the processing of a single item into two steps enables better generalization across the configurations we consider, suggesting that a single message-passing step is insufficient for both item selection and capacity updates.}
    \label{fig:one_vs_two_recon_hints}
\end{figure}

\paragraph{Runtime and Memory Requirements}
Runtime and memory requirements are comparable to the ones reported for CLRS-30 benchmark algorithms.
Apart from the homogeneity, we do not modify the computational structure of the GNN layers, so we do not incur any slowdown from the GNN architecture.
The only slowdown incurred is for reconstruction, where we do twice the amount of steps a standard model would do (see \autoref{fig:one_vs_two_recon_hints}).
All experiments were conducted on a single NVIDIA A100 GPU with 40 GB of memory.
The longest single training run for any model presented in this work was 44 minutes and 56 seconds.
All models consumed less than 10 GB of memory, which allowed us to train 4 different models simultaneously.
As a result, we were able to conduct all experiments presented in the paper (across all 3 seeds) in less than 10 hours.
Inference time is negligible, measuring less than one second per instance across all configurations.

\paragraph{Adapting Tropical Attention}
To enable comparison with our model, we made modifications to the Tropical Attention repository\footnote{\url{https://github.com/Baran-phys/Tropical-Attention/}} (commit \texttt{534ac67}).
In \texttt{dataloaders.py}, the Knapsack dataset generation was modified to use uniform random values in the range $[0, 1]$ for item values instead of integer values from \texttt{value\_range} (line 264), and the type signature of \texttt{set\_knapsack\_01} was updated to accept \texttt{float} values (line 177).
A new \texttt{greedy\_decoding} method was added to \texttt{experiment.py} to convert the model's probability predictions into valid Knapsack solutions.
The \texttt{\_eval\_one\_epoch} method (\texttt{experiment.py}, line 205) was extended to integrate greedy decoding for exact match calculation.
Exact match ratio was added as an evaluation metric.
The model was trained on samples where $n = 16$ and $C \leq 16$, and tested on the same size combinations as in our experiment.

\section{NAR Construction Implementation}
\begin{figure}[h!]
    \centering
    \includegraphics[width=0.95\textwidth]{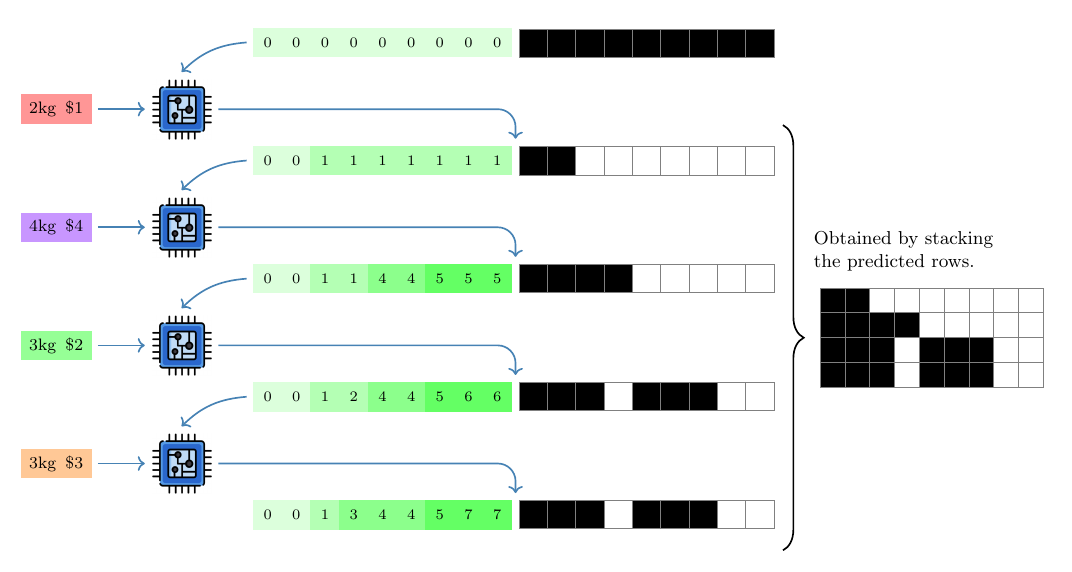}
    \caption{Visualization of the NAR construction process for the Knapsack problem.
    At each step, the next row of the DP value table and the decision table is predicted from the current latent embeddings.
    Unlike the standard CLRS-30 approach, where the output is predicted separately using additional node/edge/graph features, our model accumulates the predicted rows of the decision table, which are then passed to the NAR reconstruction.}
    \label{fig:knapsack_construction}
\end{figure}
\label{appendix:knar_construction}
\begin{lstlisting}[language=Python, caption={Python-style pseudocode for NAR construction for Knapsack. In line with CLRS-30, adding a new algorithm involves specifying its specification, implementation, and sampler.
Based on this, examples are generated and the neural model is trained.
Our only deviation from CLRS-30 is that at each step we explicitly provide the current item as input. For simplicity, this is implemented via hints (marked with \texttt{"RNN input"}), which are ignored during decoding and in the loss function.}]
# Specification:
'NAR_construction': {
  'pos': (Stage.INPUT, Location.NODE, Type.SCALAR),
  'edge_length_encoding': (Stage.INPUT, Location.EDGE, Type.CATEGORICAL),
  'dp_h': (Stage.HINT, Location.NODE, Type.SCALAR),
  'decision_h': (Stage.HINT, Location.NODE, Type.MASK),
  'input_categorical_weight': (Stage.HINT, Location.GRAPH, Type.CATEGORICAL), # "RNN input"
  'input_value': (Stage.HINT, Location.GRAPH, Type.SCALAR),                   # "RNN input"
  'dummy_output': (Stage.OUTPUT, Location.GRAPH, Type.SCALAR),
},

# Implementation:
def NAR_consturction(weights, values, capacity):
  probes = probing.initialize(specs.SPECS['NAR_construction'])
  # Edge length encoding construction is omitted for brevity, see Appendix F.
  probing.push(
    probes,
    specs.Stage.INPUT,
    next_probe={
      'pos': np.arange(capacity + 1) * 1.0 / (capacity + 1),
      'edge_length_encoding': edge_length_encoding,
    })

  # Capacity is implicitly defined through the number of nodes in the graph.
  dp = np.zeros(capacity + 1)
  decision_h = np.zeros(capacity + 1)

  for i in range(num_items + 1):
    probing.push(
      probes,
      specs.Stage.HINT,
      next_probe={
        'dp_h': dp.copy(),
        'decision_h': decision_h.copy(),
        # The following two are representetd as hints for implementation purposes,
        # but they are handled as inputs, because the true values are given in each
        # step and no prediction is done for them.
        'input_categorical_weight': to_categorical(weights[i] if i < num_items else 0),
        'input_value': values[i] if i < num_items else 0,
      })
    decision_h = np.zeros(capacity + 1)
    if i < num_items:
      for j in range(capacity, weights[i] - 1, -1):
        if dp[j] < dp[j - weights[i]] + values[i]:
          decision_h[j] = 1
          dp[j] = dp[j - weights[i]] + values[i]

  # Dummy output omitted for brevity.

  probing.finalize(probes)
  return 0, probes
\end{lstlisting}

\section{NAR Reconstruction Implementation}
\label{appendix:knar_reconstruction}
\begin{figure}[h!]
    \centering
    \includegraphics[width=0.95\textwidth]{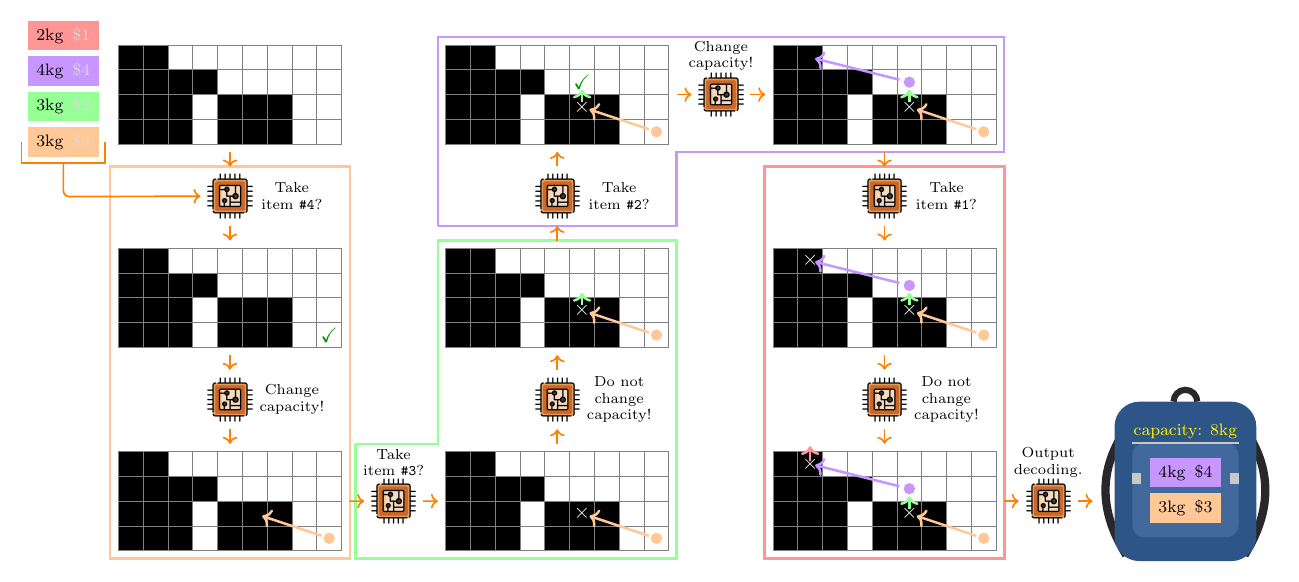}
    \caption{Visualization of the NAR reconstruction process for the Knapsack problem.
    Using the predicted $\mathrm{decision}$ table and item weights (see \autoref{fig:with_vs_without_item_values}), the model simulates item selection and traversal of the table, i.e. 2 steps per item ($2n$ in total).
    After the table is traversed, the probability of each item being part of the solution is predicted.
    While item selection and capacity updates could be merged into a single step, keeping them separate has shown better generalization (see \autoref{fig:one_vs_two_recon_hints}).
    }
\end{figure}
\begin{lstlisting}[language=Python, caption={Python-style pseudocode for NAR reconstruction for Knapsack. The hints discussed in the main text (\S2, p.\ 3) are presented in L7-L14. Out of those, the only less standard is \texttt{alteration\_type}. It is similar to the \texttt{phase} hint found in the other algorithms implemented in CLRS-30, with the exception that \texttt{alteration\_type} is \emph{always} a sequence of alternating zeros and ones. Contrary to construction, where output was obtained from stacked hints, here it is decoded from the final latent embeddings, as standard for CLRS-30.}]
# Specification:
'NAR_reconstruction': {
  'pos': (Stage.INPUT, Location.NODE, Type.SCALAR),
  'node_type': (Stage.INPUT, Location.NODE, Type.MASK),
  'categorical_weight': (Stage.INPUT, Location.NODE, Type.CATEGORICAL),
  'edge_length_encoding': (Stage.INPUT, Location.EDGE, Type.CATEGORICAL),
  'predicted_decision_table': (Stage.INPUT, Location.EDGE, Type.MASK), # Soft MASK
  'node_idx': (Stage.HINT, Location.NODE, Type.MASK_ONE),              # The item currently being processed.
  'capacity_pointer': (Stage.HINT, Location.NODE, Type.MASK_ONE),      # The remaining capacity corresponds to one
                                                                       # of our capacity nodes.
  'alternation_type': (Stage.HINT, Location.GRAPH, Type.MASK),         # Item taking step or capacity decrease step?
                                                                       # Check alternation_type.
  'take_curr_item': (Stage.HINT, Location.GRAPH, Type.MASK),           # Whether the current item is part of the final solution.
  'selected_h': (Stage.HINT, Location.NODE, Type.MASK),                # The selected items so far.
  'selected': (Stage.OUTPUT, Location.NODE, Type.MASK),                # The final solution.
}


# Implementation:
def NAR_reconsturction(weights, capacity, predicted_decision_table):
  # Predicted_decision_table are probabilities calucluated by the NAR construction model.
  # These probabilities are assigned to the edges connecting the item and capacity nodes.
  # The other edges are assigned a probability of 0.

  num_items = len(weights)

  dp = np.zeros((num_items, capacity + 1))
  decision = np.zeros((num_items, capacity + 1))
  # Calculate true dp and decision tables as in NAR_construction, omitted for brevity.

  # Edge length encoding construction is omitted for brevity, see Appendix F.
  # Capacity is implicitly defined through the number of the capacity nodes in the graph.
  probing.push(
      probes,
      specs.Stage.INPUT,
      next_probe={
        'node_type': np.concatenate([np.zeros(num_items), np.ones(capacity + 1)]),
        'pos': np.concatenate([np.arange(num_items) * 1.0 / num_items, np.arange(capacity + 1) * 1.0 / (capacity + 1)]),
        'categorical_weight': to_categorical(np.concatenate([np.array(weights), np.zeros(capacity + 1)])),
        'edge_length_encoding': edge_length_encoding,
        'predicted_decision_table': predicted_decision_table,
      })

  selected_items = np.zeros(num_items)
  curr_capacity_pointer = capacity

  for i in range(num_items, -1, -1):  # -1 = initialization
    for phase in (["init"] if i == num_items else ["taking_item", "moving_capacity_pointer"]):
      if phase == "init":
        node_idx, alternation_type, take_curr_item = 0, 0, 0
      elif phase == "taking_item":
        node_idx, alternation_type, take_curr_item = i, 1, decision[i, curr_capacity_pointer]
      else:
        node_idx, alternation_type, take_curr_item = i, 0, 0

      probing.push(
        probes,
        specs.Stage.HINT,
        next_probe={
          'node_idx': probing.mask_one(node_idx, num_items + capacity + 1),
          'capacity_pointer': probing.mask_one(num_items + curr_capacity_pointer, num_items + capacity + 1),
          'alternation_type': alternation_type,
          'take_curr_item': take_curr_item,
          'selected_h': np.concatenate([selected_items, -1 * np.ones(capacity + 1)]),
        })

      if phase == "taking_item" and take_curr_item == 1:
        curr_position -= weights[i] # Move capacity pointer.
        selected_items[i] = 1       # Add item to selected items.


  probing.push(
      probes,
      specs.Stage.OUTPUT,
      next_probe={
        'selected': np.concatenate([selected_items, -1 * np.ones(capacity + 1)]),
      })

  probing.finalize(probes)
  return selected, probes
\end{lstlisting}

\begin{figure}[h!]
    \centering
    \includegraphics[width=0.95\textwidth]{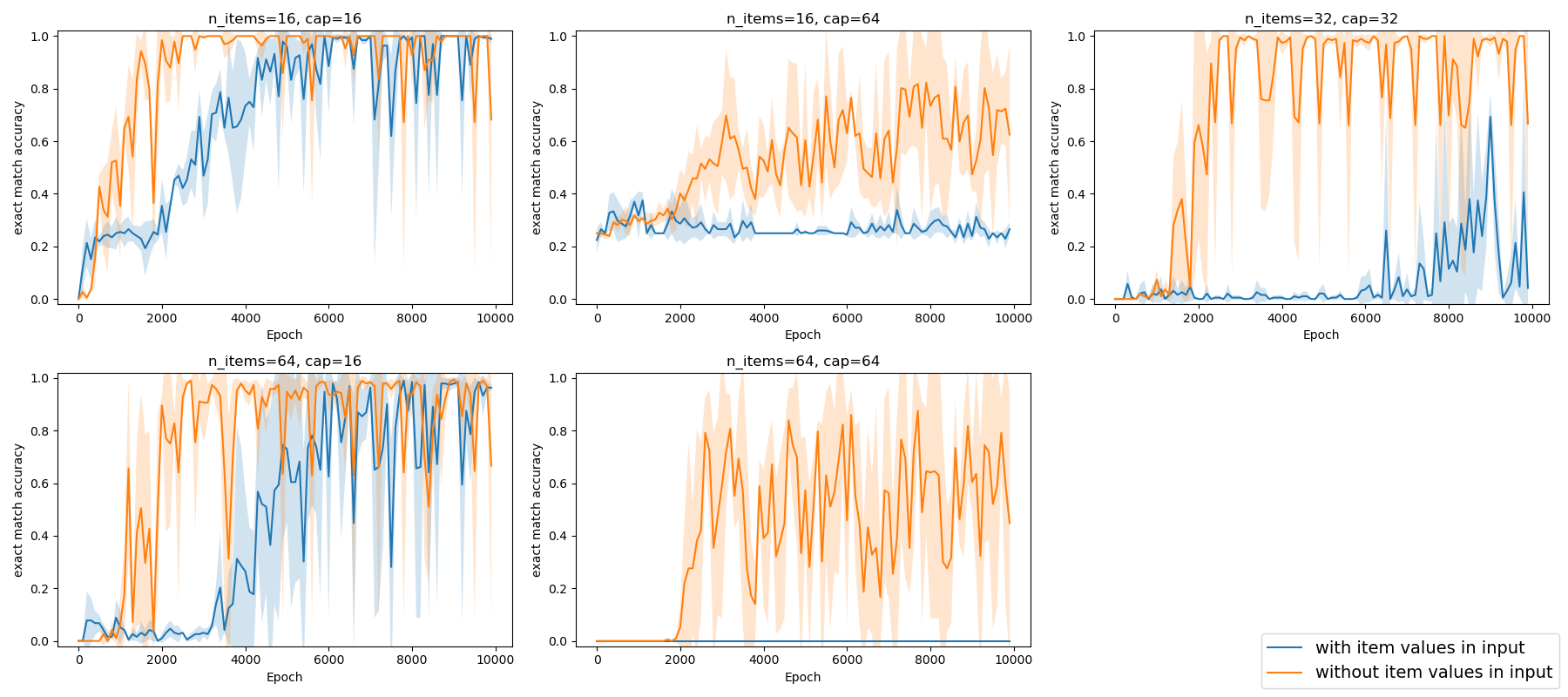}
    \caption{Comparison of NAR reconstruction performance with and without item values in the input, given the true $\mathrm{decision}$ table during both training and inference.
    Having the items' values in the input prevents the NAR reconstruction model from generalizing to larger instances.}
    \label{fig:with_vs_without_item_values}
\end{figure}

\section{Edge Length Encoding}
\label{appendix:edge_positional_encoding}
As shown in \autoref{fig:edge_positional_encoding_example}, edge length encoding for the NAR construction model assigns to each graph edge $e_{ij} \in \gE$ a categorical feature of size $M=10$ that represents the absolute difference between the capacities associated with vertices $u_i$ and $u_j$, specifically $|i-j|$. Since these are categorical features, all differences greater than or equal to $M-1$ are treated as equivalent. For the NAR reconstruction model, edge length encoding follows a similar approach, but only considers edges between the $C+1$ capacity vertices. The remaining $(C + 1 + n)^2 - (C + 1)^2$ edges are assigned the same categorical label as distances greater than or equal to $M-1$.

We further elaborate on this empirically important inductive bias by explaining its motivation.
As an example, we use the construction model, which has $C+1$ nodes corresponding to capacities $0, 1, \ldots, C$.
A similar argument can be straightforwardly applied to the reconstruction model.
Recall the DP formula $\mathrm{dp}_{i, c}=\max(\mathrm{dp}_{i-1, c}, \mathrm{dp}_{i-1, c-w_i} + v_i)$.
Observe that at the $i$-th step, which corresponds to item $i$, the following holds for every node $c$: the DP scalar feature associated with node $c$ must either remain unchanged or be updated based on the value of the DP scalar feature associated with node $c'=c-w_i$, where $c \geq w_i$.
We see that $c-c'=w_i$, meaning that for all nodes $c$, the distance to their corresponding candidate node $c'$ is equal to the item weight $w_i$.
NAR processors are based on GNNs where, along every edge $(i, j)$, a message from node $i$ to node $j$, $m_{ij}$, is computed based on the edge embedding and the embeddings of $i$ and $j$, and these messages are then aggregated across all neighboring nodes \cite{velivckovic2022clrs}.
As it is currently implemented, the CLRS-30 employs a positional encoding scheme with \textit{scalar real} values in $[0, 1]$.
As a result, the model cannot infer if either of the two nodes is a candidate for the other.
The introduction of edge length encoding, which assigns categorical features to edges corresponding to their length (i.e., the absolute difference in capacities), means that, to a large extent, computing the message $m_{ij}$ boils down to checking whether the corresponding distance equals $w_i$ or not.
Note that our use of the absolute difference $|c-c'|$ in the definition of edge length encoding reduces the number of required categories, which is possible because candidates $c' < c$ are easily identified based on the existing scalar node positional encoding.
The cutoff $M$ is used because it is necessary when employing categorical features, and its value is adjusted to the maximum considered item weight.
It is important to note that this argumentation is not limited to the Knapsack problem but can be applied to all pseudo-polynomial DP algorithms.
Specifically, the same alignment between edge length encoding and the structure of the pseudo-polynomial DP relation holds for other problems: in Subset Sum and Partition problems, the alignment corresponds to differences of possible sums, while in Minimum Coin Exchange, it corresponds to differences of possible target amounts, to name a few.

\section{Homogeneous NAR Construction}
\label{appendix:homogeneity}

\begin{figure}[h!]
    \centering
    \begin{minipage}{0.27\textwidth}
        \centering
        \includegraphics[width=\textwidth]{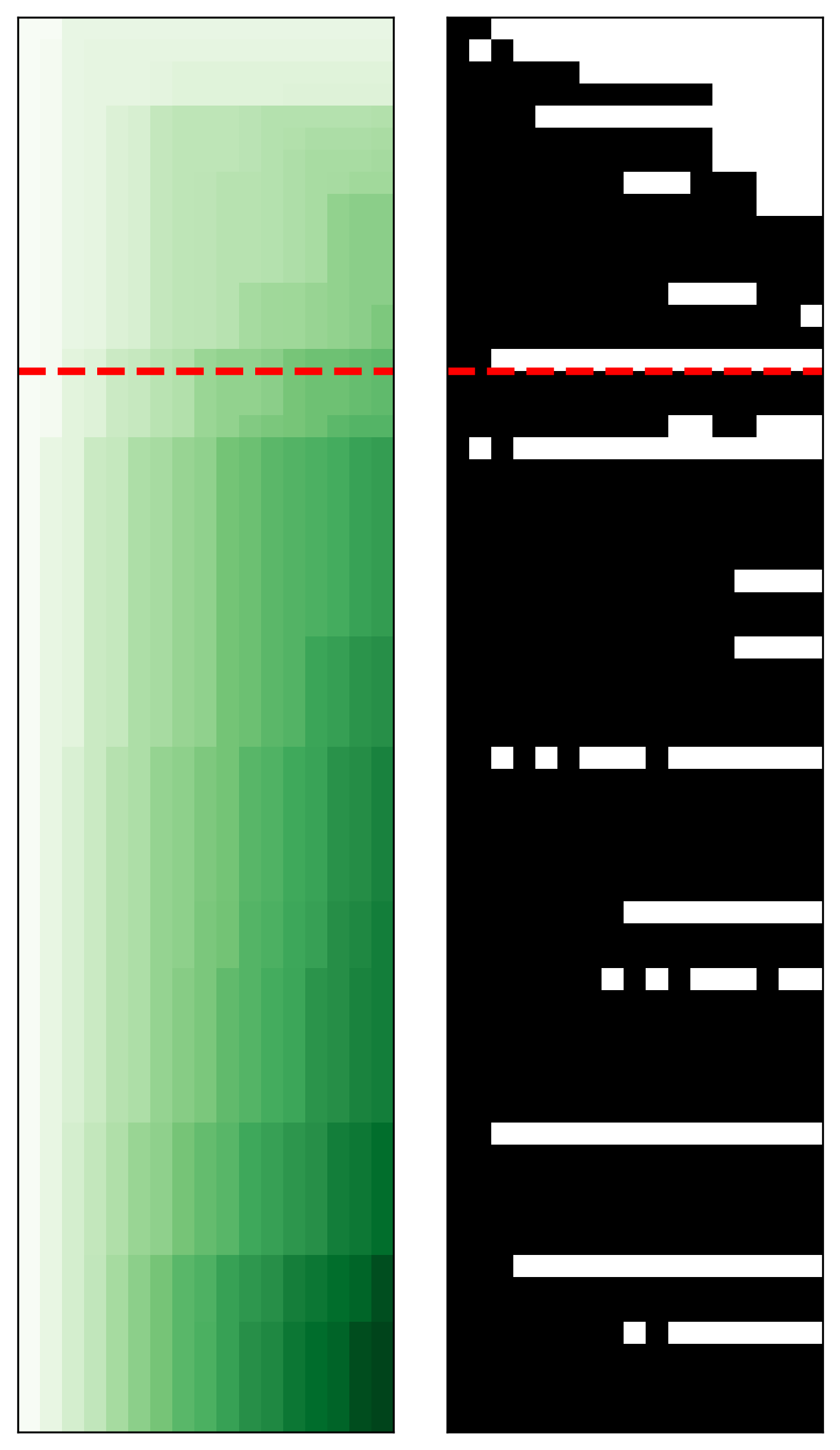}
        \subcaption{True DP value and decision table}
    \end{minipage}\hspace{0.08\textwidth}
    \begin{minipage}{0.27\textwidth}
        \centering
        \includegraphics[width=\textwidth]{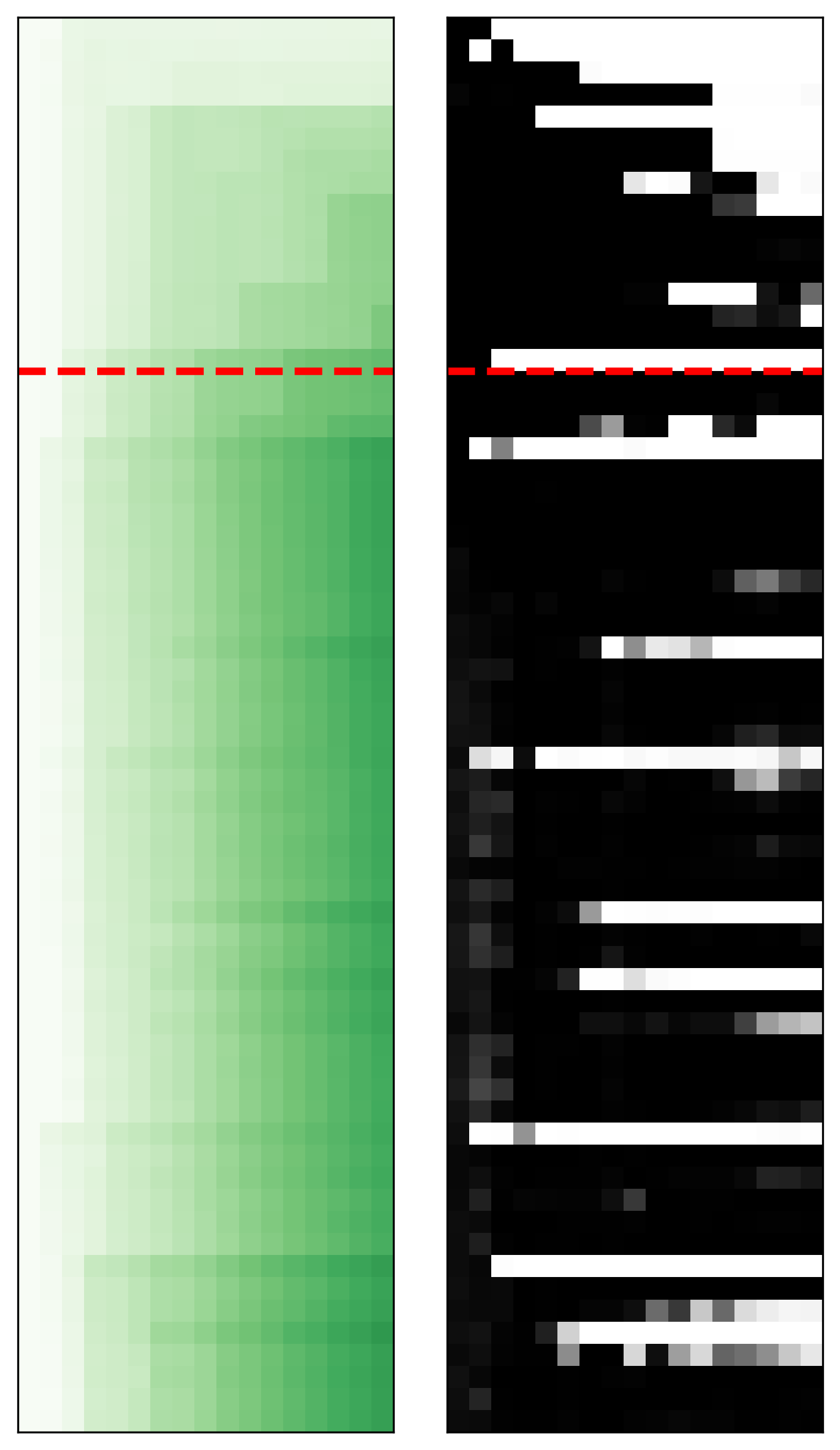}
        \subcaption{Tables predicted by regular NAR construction}
    \end{minipage}\hspace{0.08\textwidth}
    \begin{minipage}{0.27\textwidth}
        \centering
        \includegraphics[width=\textwidth]{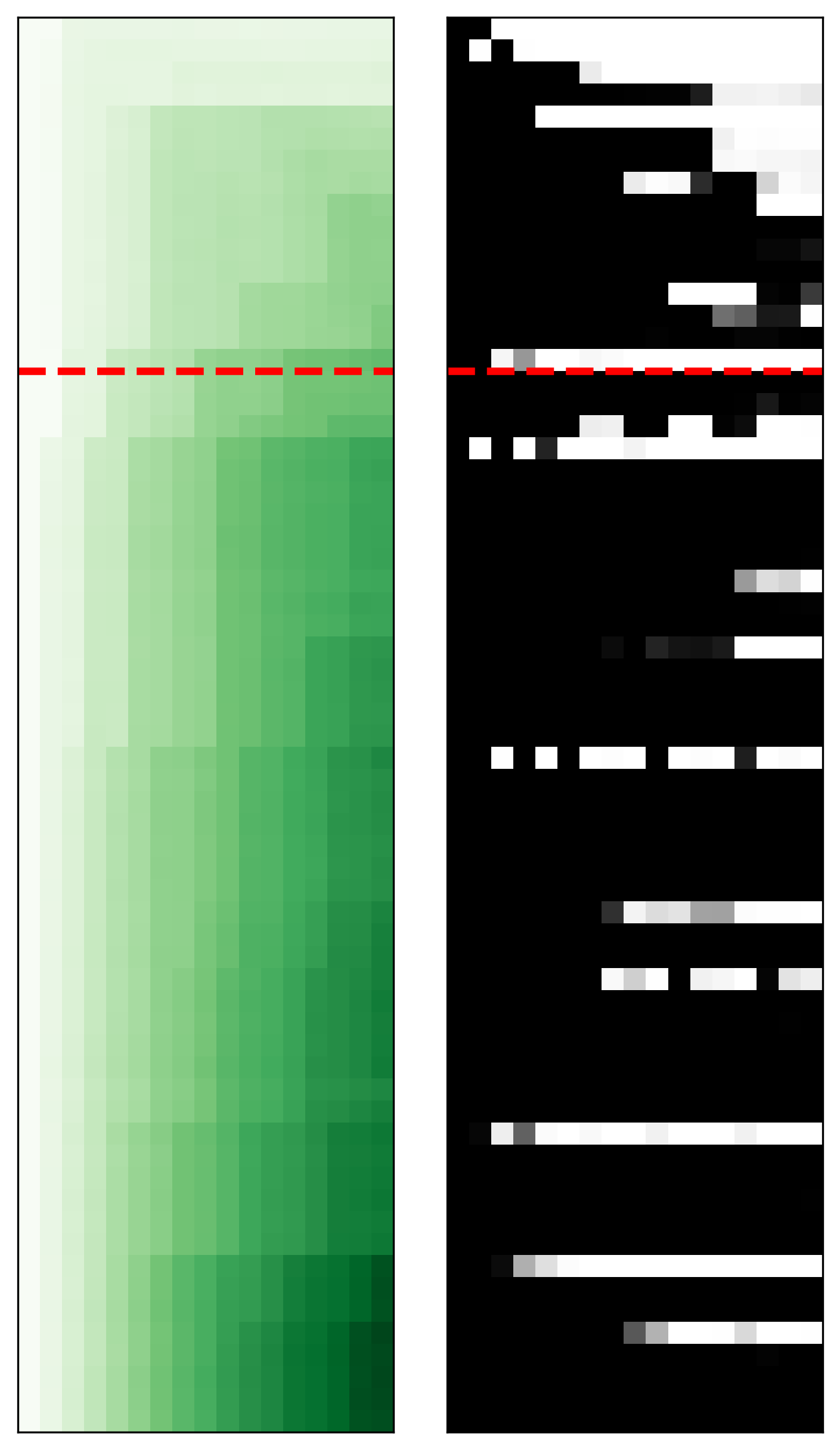}
        \subcaption{Tables predicted by homogeneous NAR construction}
    \end{minipage}
    \caption{Comparison of DP value and decision tables for $n=64$, $C=16$, examining the effect of the NAR construction model homogeneity. The lighter color in the bottom-right of the regular processor's scalar predictions indicates under-prediction of out-of-distribution values, which is corrected in the homogeneous model.}
    \label{fig:dp_decision_matrices}
\end{figure}

\begin{figure}[h!]
    \centering
    \begin{minipage}{0.48\textwidth}
        \centering
        \includegraphics[width=\textwidth]{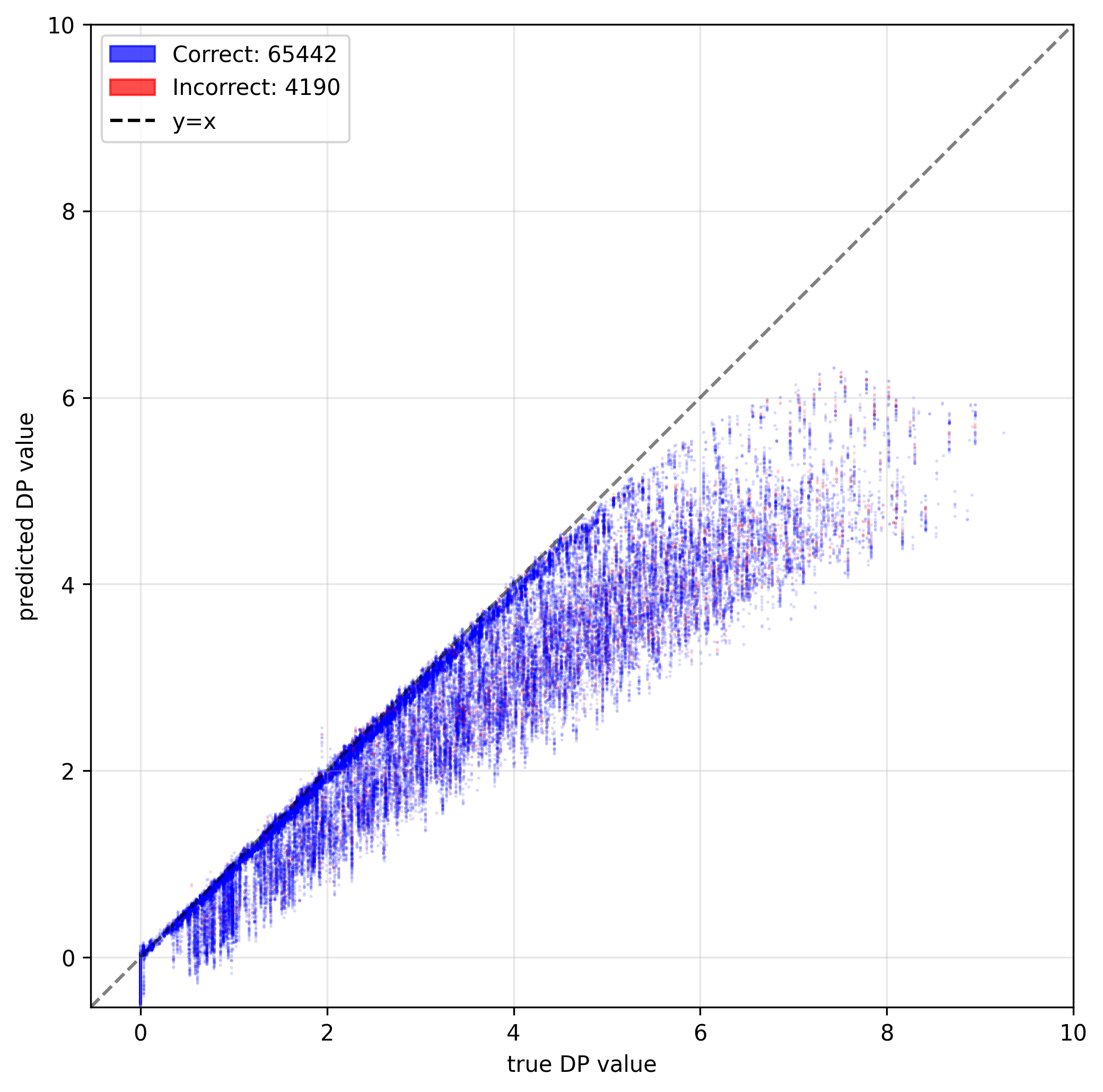}
        \subcaption{Regular NAR construction}
        \label{fig:homogeneity_correlation_regular}
    \end{minipage}\hfill
    \begin{minipage}{0.48\textwidth}
        \centering
        \includegraphics[width=\textwidth]{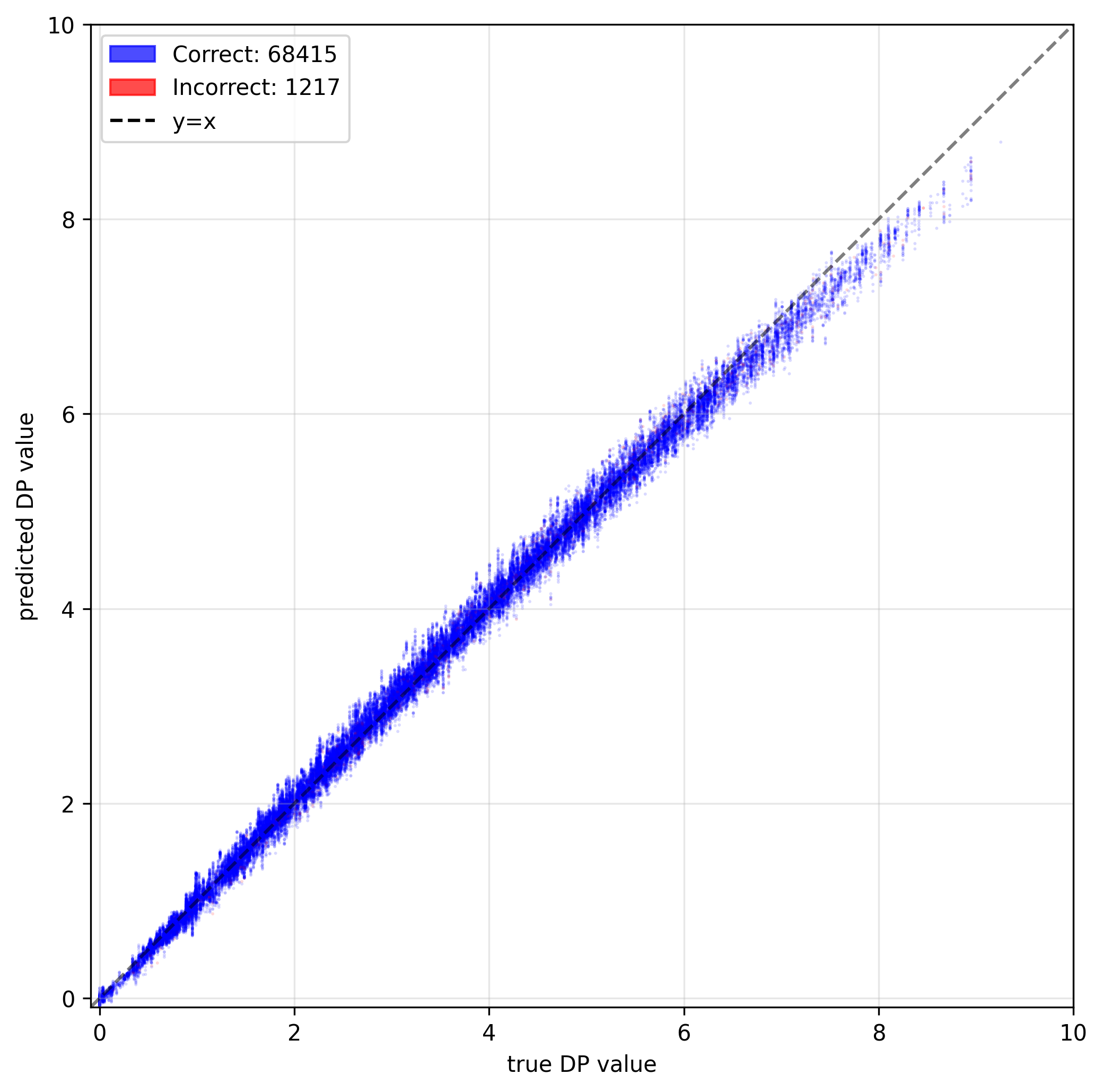}
        \subcaption{Homogeneous NAR construction}
        \label{fig:homogeneity_correlation_homo}
    \end{minipage}
    \caption{Correlation between true and predicted DP values for (a) regular and (b) homogeneous NAR construction models, considering all examples for $n=64$, $C=16$. \textcolor{blue}{Blue} dots represent elements of the DP table whose corresponding decision is correctly determined, while \textcolor{red}{red} dots represent those where it is not.}
    \label{fig:homogeneity_correlation}
\end{figure}

In \autoref{fig:dp_decision_matrices}, we observe that for the regular processor the bottom-right
scalar predictions appear lighter than expected, indicating for out-of-distribution the model is predicting lower DP scalar values than the ground-truth ones.
A similar phenomenon was reported by \citet{mirjaniclatent} in the context of the Bellman-Ford algorithm, where it was suggested to scale the embeddings by a constant factor $c < 1$ at every message-passing step.
Differently from them, we address this problem by \emph{constraining our processor to be homogeneous}~\cite{tang2020towards}, which enforces the
model to be invariant with respect to the scale ($f(\alpha x)=\alpha f(x), \alpha>0$) of the item values. Concretely, this entails
removing bias terms from the network, as well as eliminating layer normalisations \citep{ba2016layernormalization} and gating mechanisms \citep[p.5]{ibarz2022generalist}.
Additionally, in each step, it is necessary to encode only homogeneous $\mathrm{dp}_{i, :}$ hints, while applying the \textit{decode-only} hint mode for other hints.
As shown in \autoref{fig:homogeneity_correlation}, these modifications improve the
correlation between true and predicted DP values, and lead to better generalization of the NAR
construction model (\autoref{tab:homogeneity_construction}), which enables significantly better overall NAR model performance compared to the no-hint baseline.

\begin{table}[h!]\centering
	\caption{\textcolor{Orange}{In-} and \textcolor{DarkOrchid}{out-} of distribution performance comparison of regular vs. homogeneous NAR construction models, tested on original and 10$\times$ scaled item values. From the 10$\times$ results we observe that the homogeneous model is indeed invariant to item value scaling, while the poor performance of the regular model in this case further confirms its inability to generalize to larger scalar values.}
	\label{tab:homogeneity_construction}
	\begin{tabular}{c cccc}
	\toprule
	\multirow{2}{*}{mult.} & \multirow{2}{*}{$n$} & \multirow{2}{*}{$C$} & \multicolumn{2}{c}{micro-F1} \\
	\cmidrule(lr){4-5}
	& & & reg. c. & homo. c. \\
	\midrule
	\multirow{5}{*}{1} & \textcolor{Orange}{16} & \textcolor{Orange}{16} & $\mathbf{0.991}_{\scriptscriptstyle\pm0.001}$ & $0.989_{\scriptscriptstyle\pm0.002}$ \\
	\cmidrule(lr){2-5}
	& \textcolor{DarkOrchid}{16} & \textcolor{DarkOrchid}{64} & $\mathbf{0.978}_{\scriptscriptstyle\pm0.005}$ & $0.976_{\scriptscriptstyle\pm0.013}$ \\
	& \textcolor{DarkOrchid}{32} & \textcolor{DarkOrchid}{32} & $0.958_{\scriptscriptstyle\pm0.008}$ & $\mathbf{0.972}_{\scriptscriptstyle\pm0.013}$ \\
	& \textcolor{DarkOrchid}{64} & \textcolor{DarkOrchid}{16} & $0.909_{\scriptscriptstyle\pm0.030}$ & $\mathbf{0.972}_{\scriptscriptstyle\pm0.011}$ \\
	& \textcolor{DarkOrchid}{64} & \textcolor{DarkOrchid}{64} & $0.847_{\scriptscriptstyle\pm0.033}$ & $\mathbf{0.861}_{\scriptscriptstyle\pm0.105}$ \\
	\midrule
	\multirow{5}{*}{10} & \textcolor{DarkOrchid}{16} & \textcolor{DarkOrchid}{16} & $0.460_{\scriptscriptstyle\pm0.182}$ & $\mathbf{0.989}_{\scriptscriptstyle\pm0.002}$ \\
	& \textcolor{DarkOrchid}{16} & \textcolor{DarkOrchid}{64} & $0.646_{\scriptscriptstyle\pm0.255}$ & $\mathbf{0.976}_{\scriptscriptstyle\pm0.013}$ \\
	& \textcolor{DarkOrchid}{32} & \textcolor{DarkOrchid}{32} & $0.436_{\scriptscriptstyle\pm0.157}$ & $\mathbf{0.971}_{\scriptscriptstyle\pm0.014}$ \\
	& \textcolor{DarkOrchid}{64} & \textcolor{DarkOrchid}{16} & $0.263_{\scriptscriptstyle\pm0.066}$ & $\mathbf{0.972}_{\scriptscriptstyle\pm0.010}$ \\
	& \textcolor{DarkOrchid}{64} & \textcolor{DarkOrchid}{64} & $0.423_{\scriptscriptstyle\pm0.121}$ & $\mathbf{0.861}_{\scriptscriptstyle\pm0.104}$ \\
	\bottomrule
	\end{tabular}
\end{table}

\clearpage
\section{Deterministic Reconstruction Implementation}
\label{appendix:deterministic_reconstruction}
\begin{lstlisting}[language=Python, caption=Python implementation of the deterministic reconstruction algorithm., label=lst:deterministic_reconstruction]
def deterministic_reconstruction(weights, capacity, decision):
    n = len(weights)
    soft_selected = numpy.zeros(n)
    dp_prob = numpy.zeros((n, capacity + 1))
    for i in range(n - 1, -1, -1):
        for j in range(capacity, 0, -1):
            if i == n - 1:
                if j == capacity:
                    dp_prob[i, j] = 1
                continue
            dp_prob[i, j] = (1 - decision[i + 1, j]) * dp_prob[i + 1, j]
            if j + weights[i + 1] <= capacity:
                dp_prob[i, j] += decision[i + 1, j + weights[i + 1]] * dp_prob[i + 1, j + weights[i + 1]]
        for j in range(weights[i], capacity + 1):
            soft_selected[i] += dp_prob[i, j] * decision[i, j]
    return soft_selected
\end{lstlisting}
Note that in our two-phase NAR approach, the reconstruction model is always trained on the true discrete steps of classical backtracking, regardless of the probabilistic decision table predicted by the NAR construction model.
The classical reconstruction procedure relies on \emph{discrete} dynamic programming: at each state $(i,j)$ one either includes or excludes item $i$, producing a single path and a final 0--1 solution vector. This process is not differentiable, as it involves the discrete $\arg\max$ operator.

We introduce a \emph{deterministic relaxation} of this process (see \autoref{lst:deterministic_reconstruction}). Instead of hard choices, we employ a table $\texttt{decision}[i,j] \in [0,1]$, which represents the ``soft'' probability of including item $i$ given remaining capacity $j$. Using these probabilities, we recursively compute a distribution of probability mass $\texttt{dp\_prob}[i,j]$, describing the probability of reaching state $(i,j)$ during reconstruction. At each branching step, the mass is split proportionally according to $\texttt{decision}$ (for inclusion) and $1 - \texttt{decision}$ (for exclusion). In this way, the algorithm maintains a superposition of all feasible backtracking paths, rather than committing to a single one.
Finally, the expected inclusion of item $i$ is obtained as
\[
\texttt{soft\_selected}[i] \;=\; \sum_{j=0}^{C} \texttt{dp\_prob}[i,j] \cdot \texttt{decision}[i,j].
\]
This produces values in $[0,1]$, which can be interpreted as a continuous relaxation of binary inclusion decisions.
The procedure is differentiable since the recursion involves only additions and multiplications of continuous variables. No discrete operators (such as $\arg\max$ or indicator functions) appear, ensuring that gradients can propagate through the entire reconstruction phase.
\end{document}